\documentclass[10pt,twocolumn,letterpaper]{article}

\usepackage{cvpr}
\usepackage{times}
\usepackage{epsfig}
\usepackage{graphicx}
\usepackage{amsmath}
\usepackage{amssymb}
\usepackage{algorithm}
\usepackage{algorithmic}
\usepackage{amsmath}
\usepackage{mathrsfs}
\usepackage{multirow}
\usepackage{color}
\usepackage{bm}
\usepackage{float}
\usepackage{booktabs}       
\usepackage{microtype}
\usepackage{subfigure}
\usepackage{xcolor}
\usepackage{colortbl}

\usepackage[breaklinks=true,bookmarks=false]{hyperref}

\cvprfinalcopy 

\def\cvprPaperID{****} 

\begin{document}

\title{Adversarial Attack across Datasets}

\author{Yunxiao Qin\\
Communication University of China\\
{\tt\small qinyunxiao@cuc.edu.cn}
\and
Yuanhao Xiong\\
UCLA\\
{\tt\small yhxiong@cs.ucla.edu}
\and
Jinfeng Yi\\
JD Technology\\
{\tt\small yijinfeng@jd.com}
\and
Liong Cao\\
Communication University of China\\
{\tt\small lihong.cao@cuc.edu.cn}
\and
Cho-Jui Hsieh\\
UCLA\\
{\tt\small chohsieh@cs.ucla.edu}
}

\maketitle
\newcommand{\tabincell}[2]{\begin{tabular}{@{}#1@{}}#2\end{tabular}}  

\begin{abstract}
	Existing transfer attack methods commonly assume that the attacker knows the training set (e.g., the label set, the input size) of the black-box victim models, which is usually unrealistic because in some cases the attacker cannot know this information. In this paper, we define a Generalized Transferable Attack (GTA) problem where the attacker doesn't know this information and is acquired to attack any randomly encountered images that may come from unknown datasets. To solve the GTA problem, we propose a novel Image Classification Eraser (ICE) that trains a particular attacker to erase classification information of any images from arbitrary datasets. Experiments on several datasets demonstrate that ICE greatly outperforms existing transfer attacks on GTA, and show that ICE uses similar texture-like noises to perturb different images from different datasets. Moreover, fast fourier transformation analysis indicates that the main components in each ICE noise are three sine waves for the R, G, and B image channels. Inspired by this interesting finding, we then design a novel Sine Attack (SA) method to optimize the three sine waves. Experiments show that SA performs comparably to ICE, indicating that the three sine waves are effective and enough to break DNNs under the GTA setting.
\end{abstract}

\section{Introduction}
\label{sec:introduction}
The fast development of Deep neural networks (DNNs) has promoted many artificial intelligence fields ranging from image classification\cite{he2016deep} to natural language translation\cite{zhang2019bridging} and self-driving cars\cite{prakash2021multi}.
However, lots of studies~\cite{szegedy2014intriguing,goodfellow2014explaining,carlini2017towards,lin2020nesterov} demonstrated that by adding human imperceptible adversarial perturbations to clean input data, even well-trained DNNs can be fooled with high probabilities. 
This phenomenon indicates that DNNs are vulnerable to adversarial attacks and reveals their robustness issue.

To evaluate the robustness of trained DNNs, lots of adversarial attack generation methods were proposed in recent years.
According to how many victim model information the attacker can access, existing adversarial attack methods can mainly be categorized into the following three main stream settings: white-box~\cite{goodfellow2014explaining,moosavi2016deepfool}, query-based black-box~\cite{brendel2018decision-based,ilyas2018black,cheng2018query}, and query-free black-box attack~\cite{liu2016delving,papernot2017practical}, which can be summarized in Table \ref{tab:problem}.
In the white-box attack setting, the attacker can access all information of the victim model to generate adversarial attacks, which is often idealistic and impractical.
Query-based black-box attack assumes that except for querying the output or prediction of the victim model, all the other information of the victim model is hidden from the attacker.
Among the three settings, query-free black-box attack is the most challenging one because as Table \ref{tab:problem} shows, even querying the victim model is also forbidden to the attacker.

\begin{table*}[]
	\centering
	\footnotesize 
	\caption{The information that the attack methods can access.
	}
	\begin{tabular}{cc}
		\toprule
		Attack Setting & Information of the target model that can be accessed \\
		\midrule
		White-Box & \tabincell{c}{All information (network architecture, network weight, gradient, score, \\ prediction, input-resolution,  output-dimension, label-set, \emph{etc.})} \\
		\midrule
		Query-based Black-Box & \tabincell{c}{ Limited information (prediction, score, input-resolution, label-set) } \\
		\midrule
		Query-free Black-Box & Limited information (input-resolution, label-set) \\
		\midrule
		Generalized Transferable Attack & Non information (/) \\
		\bottomrule
	\end{tabular}
	\label{tab:problem}
\end{table*}

\begin{figure*}[h]
	\centering
	\includegraphics[width=0.95\textwidth]{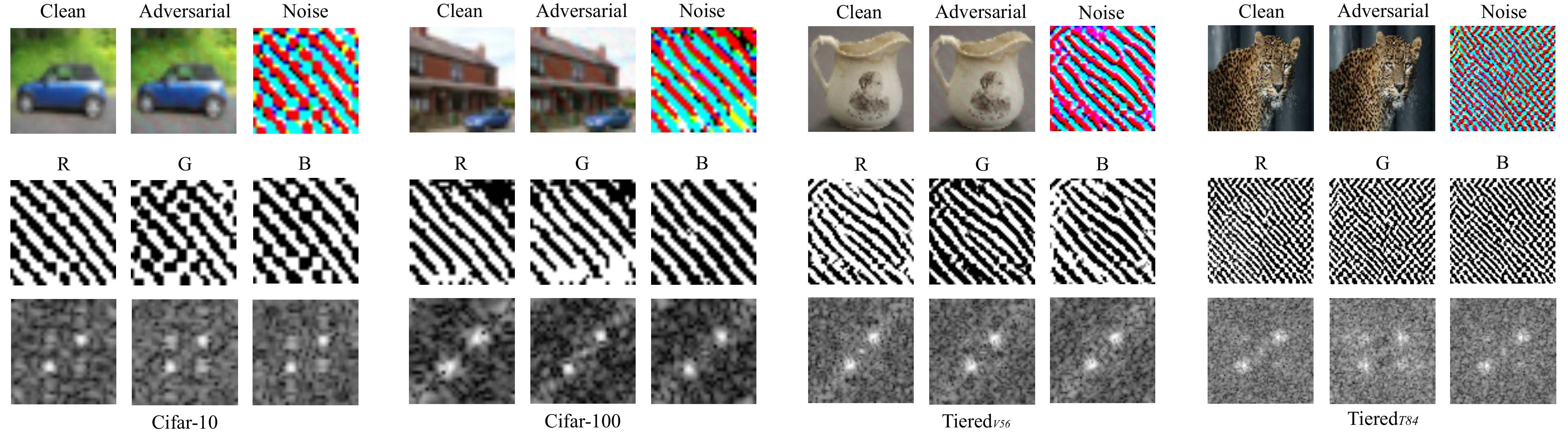}
	\caption{
		For each of the four datasets Cifar-10, Cifar-100, Tiered$_{V56}$, and Tiered$_{T84}$, the first raw shows the sampled clean image, the corresponding adversarial image and ICE noise.
		The second raw: the three channels of the ICE noise.
		The third raw: the spectrum diagram of each noise channel.
		Note that all the original noises are constrained into the range of $-\epsilon$ to $\epsilon$, and each spectrum diagram are calculated on the original noise span, but for visualization, we normalize the visualized noises into the range of 0 to 255.
		Resolutions of Cifar-10, Cifar-100, Tiered$_{V56}$, and Tiered$_{T84}$ are 32$\times$32, 32$\times$32, 56$\times$56, and 84$\times$84, respectively.
		We resize all resulting adversarial images and ICE noises into a consistent larger resolution for clear visualization.
		Tiered$_{V56}$ and Tiered$_{T84}$ are two sub-datasets of TieredImageNet and contain no overlapping image categories.
		Section~\ref{sec:experiment} will detail Tiered$_{V56}$ and Tiered$_{T84}$.
	}
	\label{fig:fig1}
\end{figure*}

Unfortunately, the victim model is still vulnerable due to the demonstrated transferability of adversarial examples ~\cite{wu2018understanding,wu2020boosting,naseer2019cross-domain,demontis2019why}.
The transferability-based adversarial attack methods (shorten as transfer attack in the rest of the paper) usually leverage one or a few surrogate white-box models to construct adversarial examples and transfer the obtained adversarial examples to attack the victim model without querying. 
Though previous works have demonstrated the effectiveness of transfer attacks~\cite{papernot2017practical,guo2020backpropagating} and methods have been developed to improve the transferability of adversarial examples~\cite{li2020learning,wang2021enhancing}, they usually implicitly assume that {\bf surrogate white-box models and the victim model are trained on the same dataset or are using the same label set}, and their experimental results commonly depend on this assumption.  
This means the attacker actually knows some information (e.g., {\bf input resolution} and {\bf label set}) of the victim model under the query-free black-box attack setting. 
For example, when the victim model and the victim images are from Cifar-10~\cite{krizhevsky2009learning}, they assume that the surrogate models are also trained on Cifar-10 rather than ImageNet~\cite{deng2009imagenet}. 
Despite the three main-stream attack settings, the authors of \cite{naseer2019cross-domain} proposed a more challenging cross-domain attack setting where the surrogate models and the victim model are trained on different datasets, but they still implicitly assumed that the image resolutions of victim models and the victim model are consistent.

In practice, however, the victim image (and victim model) can randomly come from any dataset with arbitrary label and  resolution, and the attacker won't know which dataset is being used in advance.  It is also impractical to retrain new surrogate models for each new dataset. 
Therefore, we need to assume the surrogate models and the victim model are trained on different datasets with different label sets and different image resolutions.  
We denote this more practical setting as \textbf{g}eneralized \textbf{t}ransferable \textbf{a}ttack (\textbf{GTA}) because the attacker needs to be generalized to attack \textbf{images from unknown datasets} and attack \textbf{any unknown models} predicting these images. Table~\ref{tab:problem} illustrates the difference between GTA and the previous attack settings.

In this paper, we aim to investigate the robustness of DNNs under the GTA setting, and explore what kind of attacks can break DNNs under this setting.
Although most of the existing transfer attacks are not designed for attacking across datasets, with some careful modifications on the attack loss and rescaling techniques, it is possible to extend existing attacks~\cite{dong2019evading,xie2019improving,dong2018boosting} 
to this new setting (we will discuss these modifications in Sections \ref{sec:evaluate of baselines} and Appendix). Unfortunately, as will be seen in the experimental results, even with these modifications the existing transfer attacks suffer from poor attack success rates due to the mis-match of label set and input size between source and target models. 
To tackle these challenges, we sequentially propose two novel methods where the second one is highly inspired by the phenomenons exhibited by the first one.

The \textbf{first} method is inspired by meta-learning\cite{finn2017model,qin2020layer,SNAIL}.
It uses resources (\emph{i.e.}, some source datasets containing various label sets and input sizes, some white-box models trained on these source data) to meta-train~\cite{finn2017model,SNAIL,meta-teacher,liu2019metapruning} a universal surrogate model for the objective that classification information of each image from any category with any resolution can be erased by attacking the universal surrogate model. 
After meta-training the universal surrogate model, we use it to erase classification information of new images that are outside the label space and image resolution space of the source datasets.
We call the first method Image Classification Eraser (\textbf{ICE}) and will describe the detail in Section~\ref{sec:method-ICE}.

The experimental results in Section~\ref{sec:experiment} will demonstrate the effectiveness of the proposed ICE under the GTA setting.
Despite the performance, here we want to emphasize an interesting observation shown by ICE.	
To save space, we only visualize a few ICE noise examples in Fig.\ref{fig:fig1} and will visualize the noises generated by the other methods in Section \ref{sec:visualization}.
Fig.~\ref{fig:fig1} clearly shows that \textbf{ICE uses similar texture-like noise to perturb different images, even when the images are sampled from different datasets with different resolutions.}
Moreover, Section~\ref{sec:transfer noises} 
will demonstrate that the ICE noise generated for each image can be directly used to attack other images from the same or different datasets.
In other words, ICE can be viewed as a universal noise generator.
Compared with existing universal adversarial attacks\cite{moosavi2017universal,zhang2021data} who commonly focus on constructing the adversarial noise that is universal only within a single dataset, ICE is more generalizable because the generated ICE noise is universal not only within a single dataset but also across datasets and resolutions.

\begin{figure}[t]
	\centering
	\includegraphics[width=0.43\textwidth]{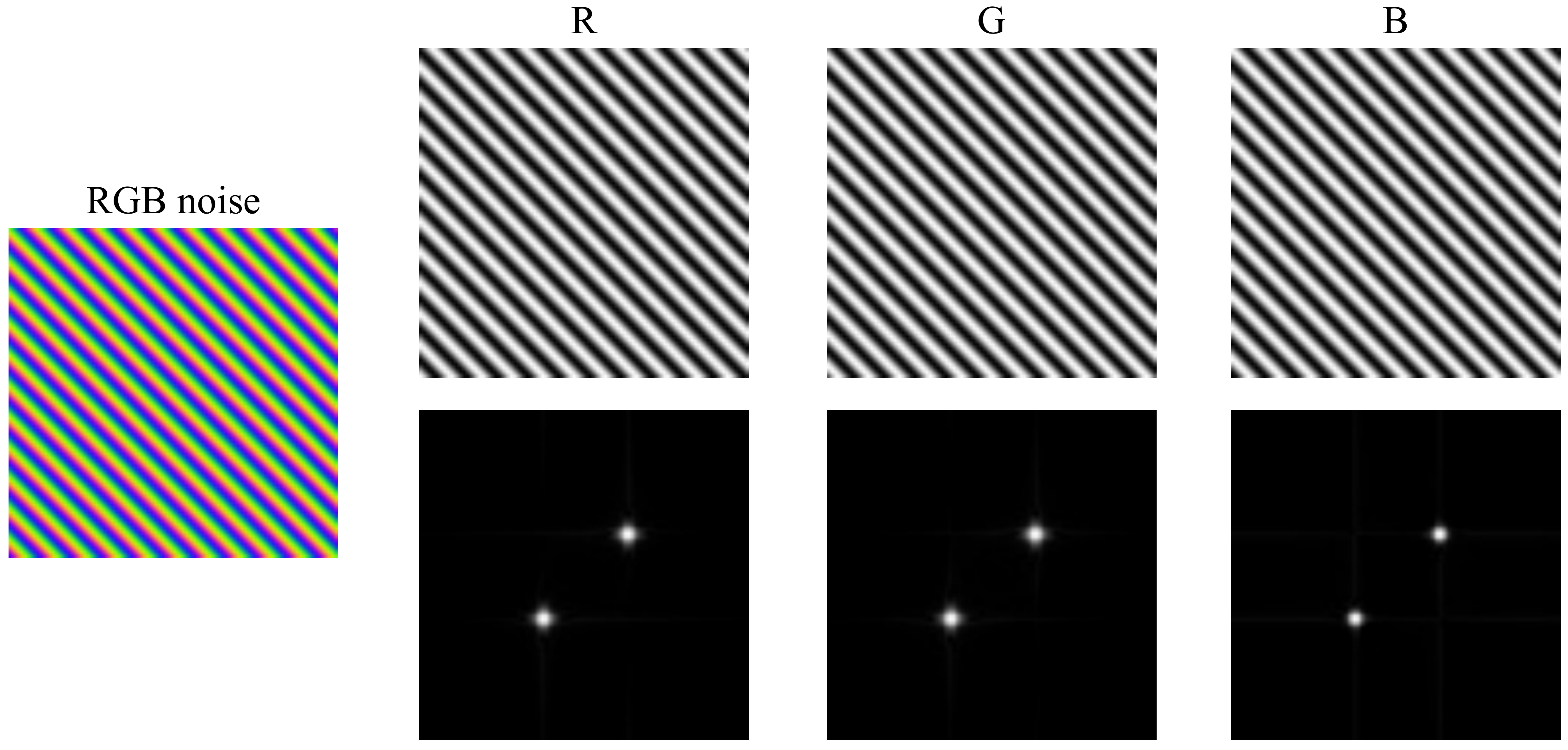}
	\caption{
		The optimized three sine waves when we use Cifar-10, Cifar-100, and Tiered$_{V56}$ as source datasets.
		The left most RGB noise is constructed by the three sine wave maps for the R, G, and B channel, respectively.
		The image under each sine wave map is the corresponding spectrum diagram.
	}
	\label{fig:fig2}
\end{figure}

As Fig.\ref{fig:fig1} shows, we also analyze noises generated by ICE via fast fourier transformation (FFT) and find that the main components in each ICE noise are three sine waves for the input R, G, and B channels.
Each sine wave is represented by the pair of the brightest points in each spectrum diagram.
Inspired by this interesting insight, we design the \textbf{second} method Sine Attack (\textbf{SA}) which directly optimizes the three sine waves for the R, G, and B channels.
Fig.\ref{fig:fig2} shows the optimized sine signals when we using Cifar-10, Cifar-100, and Tiered$_{V56}$ as source datasets.
The experiments in Section \ref{sec:experiment} demonstrate that SA performs comparably to ICE under the GTA setting, indicating that \textbf{three particular sine signals are effective adversarial signals under the GTA setting}.

The main contributions of our paper are as follows.
\begin{itemize}
	\item To the best of our knowledge, we are the first to propose the challenging and practical generalized transfer attack (GTA) setting. 
	Under GTA, the attacker is acquired to directly attack any images without knowing the image label, the dataset that the image belongs to, and the image resolution, in advance.
	\item To generate adversarial example under the GTA setting, we propose a novel meta-learning inspired method ICE. 
	Throughout experiments in Section~\ref{sec:experiment} demonstrate the effectiveness of the proposed ICE.
	\item Despite the excellent performance of ICE, we also analyze the noises generated by ICE via both experiments and FFT, and we find that ICE can be viewed as a generator for three universal sine signals for the GTA setting.
	Our finding may inspire researchers to analyze and improve the robustness of DNNs.
	\item Inspired by our finding, we propose another method SA to address the challenging GTA setting. 
	Throughout experiments in Section~\ref{sec:experiment} demonstrate that SA
	performs comparably to ICE under the GTA setting. 
	\item Furthermore, the experiments in Sections \ref{sec:transfer noises} \ref{sec:attacking CUB}, \ref{sec:attacking ImageNet}, and \ref{sec:attacking object detection} demonstrate that the ICE noise and SA noise are universal perturbations across images, datasets, resolutions, and tasks.
	For example, Section~\ref{sec:attacking object detection} shows that the SA noise constructed by three sine waves can be directly used to effectively attack object detection models.
\end{itemize}

\section{Background}
\subsection{Adversarial attack settings}
Existing adversarial attack methods~\cite{ganeshan2019fda,croce2020minimally,wu2020stronger,li2020practical,kaidi2019structured,maksym2020square,xiao2021graph,zhang2021data} can mainly be categorized into white-box, query-based black-box, and query-free black-box attacks.
Among the three kinds of attacks, white-box attack~\cite{kurakin2016adversarial} is the most effective one because all information of the target model can be leveraged to generate adversarial examples.
Query-based black-box attack assumes that some information of the target model is hidden from users and the users can only query the target model and access the hard-label or soft-label predictions.
Researchers have proposed many query-based black-box attack methods~\cite{chen2017zoo,cheng2020signopt,huang2020black,gao2020patch} and have shown that adversarial examples can still be effectively generated only based on predictions. 
The most recently developed query-based methods mainly focus on improving the querying efficiency and reducing the query counts~\cite{cheng2019improving,li2020qeba,du2020query-efficient,wang2020spanning,yuan2021meta}.
The query-free black-box attack further assumes that the target model's prediction is also hidden from users.
In this challenging situation, researchers usually generate adversarial examples via attacking surrogate models~\cite{dong2019evading,xie2019improving}.
Then, by leveraging the transferability of adversarial examples~\cite{papernot2016transferability,tramer2017ensemble,nathan2020perturbing}, we can directly use the adversarial examples to attack the target model without querying~\cite{huang2019enhancing,zhou2018transferable,Lu_2020_CVPR}. 
The level of information that the attacker can leverage reduces from white-box to query-free black-box attacks.
However, the three most popularly used attack settings still need to know some information of the victim model, which is summarized in Table \ref{tab:problem}.
For example, transfer-based attacks commonly assume the surrogate model and victim model are trained on the same dataset, which indicates that the attacker implicitly knows the dataset, input size, and label set of the victim model.

Beyond the three main stream settings, a few other attack settings have also been studied in recent works~\cite{li2020practical,nathan2020perturbing,inkawhich2021can,zhang2021data,naseer2019cross-domain}.
For instance, in no-box attack~\cite{li2020practical}, the attacker can access neither a large scale training data nor pre-trained models on it.
To attack a victim image, the no-box attacker needs firstly to collect a mini-training set that contains a small number of images and secondly to train a substitute model for generating adversarial noise on the victim image.
Due to the two steps in attacking each victim image, no-box attack is costly to attack a large amount of diverse images in practice.
The authors of \cite{inkawhich2021can} studied the problem where the white-box model and victim model share no label.
They firstly query the victim model by using the white-box dataset and obtain the mapping between white-box labels and black-box labels, and then use the white-box models and the resulting mapping to attack victim images.
Though their study is interesting, their method is sometimes unrealistic since before attacking, they have to build the mapping via extensively querying the victim model.
Universal adversarial perturbation (UAP) \cite{zhang2021data} aims to obtain a universal perturbation pattern within an image distribution so that adding the perturbation pattern to any clean image sampled from the distribution can possibly produce an adversarial example.
However, as will be demonstrated in our experiment, UAP performs not well enough to attack the images sampled from unknown distributions (datasets).
The authors of \cite{naseer2019cross-domain} proposed a cross-domain attack setting where the surrogate models and the victim model are trained on different datasets.
However, they also implicitly assumed that the resolutions of victim models and the victim model are consistent, which is still unrealistic in some challenging cases.
In this paper, 
we consider a novel and more challenging problem called generalized transferable attack (GTA) and will detail the GTA setting in Section~\ref{sec:GTA}.

\subsection{Adversarial perturbation's frequency characteristic}
Some works \cite{geirhos2019imagenet-trained,ilyas2019adversarial,das2018shield} tried to explain why adversarial perturbation exists from the perspective of frequency.
It is demonstrated that DNNs' predictions are simultaneously determined by image's low-frequency and high-frequency (including human imperceptible) information, especially the high-frequency\cite{yin2019fourier}.
For example, by using synthesized texture-shape cue conflict images, the work \cite{geirhos2019imagenet-trained} demonstrated that ImageNet CNN models are usually more sensitive to image's texture information than to shape information.
Texture is high-frequency information while shape is low-frequency.
Via experiments and analysis, the work \cite{yin2019fourier} demonstrated that instance dependent adversarial perturbations are high frequency signals and used Auto-Augmentation\cite{Cubuk_2019_CVPR} to improve DNNs' robustness towards high frequency perturbations.
Similar high-frequency characteristic was also found in universal adversarial perturbations by \cite{zhang2021universal}.
Since DNNs commonly are sensitive to high-frequency information, their predictions will possibly be determined by adversarial perturbation containing rich high-frequency information rather than the original clean image.
Though these previous works have demonstrated the high-frequency characteristic of adversarial perturbation, they didn't analyze how high the frequency should be, especially under the challenging GTA setting.
In this work, inspired by the observation shown in Fig.\ref{fig:fig1}, we propose a novel Sine Attack method which directly optimizes the parameters determining not only the frequency but also the sine wave direction and initial phase, for generating effective adversarial attacks under the GTA setting.

\section{Generalized transferable attack}
\label{sec:GTA}
GTA can be described as the following attack scene.
We have the resources of:
1) $m$ source image classification datasets denoted as $\mathcal{D}_1$, $\mathcal{D}_2$, ...,   $\mathcal{D}_m$.
Different datasets have different label sets, with potentially different label sizes and image shapes. 
2) $N_k$ trained models denoted as $\mathbf{M}_{\mathcal{D}_k}$ = \{ $\mathbf{M}_{\mathcal{D}_k}^1$, $\mathbf{M}_{\mathcal{D}_k}^2$, ..., $\mathbf{M}_{\mathcal{D}_k}^{N_k}$ \} for each source dataset $\mathcal{D}_k$.
With these resources, we can prepare an adversarial attacker.
Then, given a randomly intercepted image $x$, we are required to directly leverage the resources or the prepared attacker to disturb the image and obtain an perturbed image $\hat{x}$ so that any unknown target model $\mathbb{M}$ that predicts this image will make wrong predictions for the disturbed image, which can be formulated as $\mathbb{M}(\hat{x}) \neq \mathbb{M}(x)$.
Note that the target image is randomly intercepted so it is normal that the source datasets do not contain the image category of the target image.
Further, the resolution of the target images and target models  cannot be known in advance.

\section{Image Classification Eraser}
\label{sec:method-ICE}
GTA is a great challenging setting because we cannot know any information of the victim image and the victim model in advance but are required to directly perturb the victim image.
So how can we effectively perturb any possible victim images?
Ensemble-based attack\cite{dong2018boosting,xie2019improving}, which simply ensemble all source models and perturb the victim image by attacking all the source models with gradient-based methods (\emph{i.e.}, FGSM, PGD), is a popular direction for the transfer attack setting.
However, since source models can have different input shapes and label spaces in the GTA setting, it is nontrivial to ensemble them into a single model. Furthermore, a naive ensemble may not optimize the performance for GTA. 

We propose a novel meta-learning \cite{finn2017model,qin2020layer} inspired framework called image classification eraser (ICE) to perturb images under the GTA setting.
It firstly leverages the resources mentioned in Section~\ref{sec:GTA} to meta-train a single model for the objective that classification information of any images sampled from any categories with any kinds of resolutions can be erased by confusing the model.
This trained model can be understood as a universal surrogate model for victim target models trained on any datasets.
The model will be denoted as 
$\mathcal{U}_\theta$ in the rest of this paper, where $\theta$ is the model parameter.
After meta-training the model $\mathcal{U}_\theta$, it is expected that by confusing $\mathcal{U}_\theta$ with gradient ascent, we can obtain an effective adversarial example $\hat{x}$ for any unknown intercepted image $x$ to fool an unknown target model $\mathbb{M}$.
To meta-train the model $\mathcal{U}_\theta$ for GTA, we need to consider the following issues:

1) We cannot predict the category of a randomly encountered image $x$ in advance, so no ground-truth information can be leveraged. This means the commonly used cross-entropy loss which needs the ground-truth label cannot be directly used in GTA.
We therefore use entropy instead of cross-entropy for the attack. 
Eq.~\ref{eq:entropy} shows the formulations of cross-entropy CE$(d, y)$ and entropy $\mathcal{L}(d)$, where
$d$ is a vector that denotes the prediction distribution and $y$ is the one-hot ground-truth label.
\begin{equation}
	\left\{
	\begin{array}{lr}
		d &= \text{Softmax}(\text{logit}) \\
		\mathcal{L}(d) &= -d^T \cdot \text{log}(d) \\
		\text{CE}(d, y) &= -y^T \cdot \text{log}(d)
	\end{array}
	\right. 
	\label{eq:entropy}
\end{equation}
Entropy of any distribution denotes the degree of disorder, randomness, or uncertainty of the distribution.
Therefore, by maximizing the entropy, we can obtain a perturbation that makes the input image hard to be classified without using the ground-truth label. 
Furthermore, the entropy loss also enables us to flexibly set the output dimension of model $\mathcal{U}_\theta$ without knowing the number of categories of any dataset. 
In the experiments, we set the output dimension of $\mathcal{U}_\theta$ to 1000 by default and investigate the impact of the output dimension on its performance in Appendix.

2) 
Randomly encountered images may have diverse shapes.
For example, the image-shapes from different datasets (\emph{e.g.}, Cifar-10 and ImageNet) differ from each other.
Therefore, the model $\mathcal{U}_\theta$ must be capable of directly handling images with different shapes without resizing images.
To guarantee this point, we set the network architecture of model $\mathcal{U}_\theta$ to a fully convolutional network without flatten and fully-connected layers.
All down-sampling operations are implemented by max-pooling and average-pooling.
We will show details of the network architecture in Appendix.

\subsection{Training the universal surrogate model}
\begin{figure*}[t]
	\centering
	\includegraphics[width=0.95\textwidth]{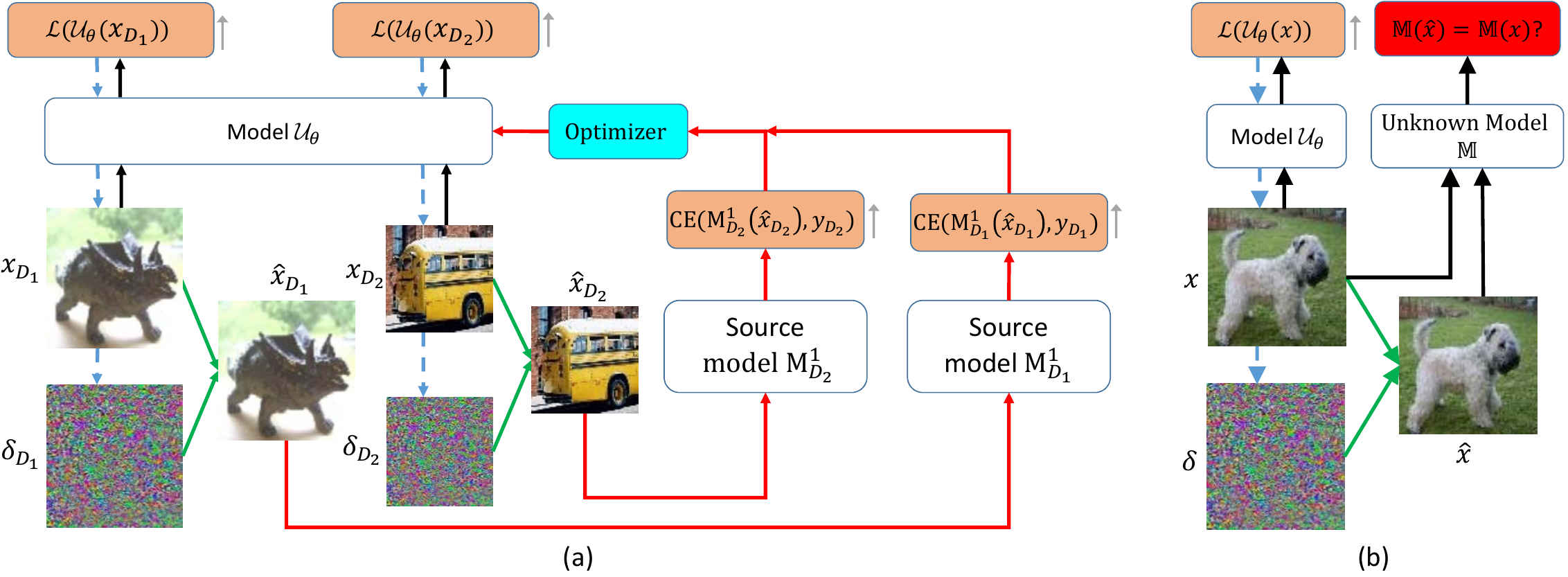}
	\caption{
		(a) The training framework of the proposed ICE.
		Different images sampled from different source datasets are used to mimic randomly encountered images and are simultaneously fed into $\mathcal{U}_\theta$. 
		By maximizing the prediction entropy of model $\mathcal{U}_\theta$ for the input images, we obtain adversarial images.
		To evaluate how confusing the generated adversarial images are, we feed them into the source models and measure the cross-entropy loss.
		Finally, 
		we optimize $\mathcal{U}_\theta$ by maximizing the source models' cross-entropy losses.
		(b) The testing pipeline of the proposed ICE.
	}
	\label{fig:framework}
\end{figure*}

To optimize the model $\mathcal{U}_\theta$, we use all source models and images to simulate unknown victim models and unknown images, respectively, and develop a novel bi-level training framework~\cite{finn2017model,liu2019metapruning,ren2018learning,liu2019darts}. 
Each bi-level training iteration contains an inner-loop and an outer-loop optimization.

\textbf{In the inner-loop}, we perturb any simulated unknown images under the GTA setting and obtain the resulting perturbed images.
Specifically, given any source dataset $D_k$ and any image $x_{\mathcal{D}_k} \in \mathcal{D}_k$, we simulate it as a randomly encountered unknown image and feed $x_{\mathcal{D}_k}$ into $\mathcal{U}_\theta$ and obtain the prediction $\mathcal{U}_\theta(x_{\mathcal{D}_k})$.
Then we generate adversarial example $\hat{x}_{\mathcal{D}_k}$ by maximizing the entropy of $\mathcal{U}_\theta(x_{\mathcal{D}_k})$.
To enable the gradient back-propagating in the bi-level optimization framework, we maximize the model $\mathcal{U}_\theta$'s prediction entropy via one-step Customized PGD (Customized FGSM)~\cite{qin2021training}. 
The reason why we conduct one-step but not multi-step attack in the inner loop will be described in Appendix.
Then, we can formulate $\hat{x}_{\mathcal{D}_k}$ as
\begin{equation}
		\left\{
		\begin{array}{lr}
			g(\theta) = \nabla_{x_{\mathcal{D}_k}}\mathcal{L}(\mathcal{U}_\theta(x_{\mathcal{D}_k})) \\
			\hat{x}_{\mathcal{D}_k} \!\! = \! \text{Clip}\Big(x_{\mathcal{D}_k} \!\!+ \epsilon_c \! \cdot\! \big (\gamma_1 \!\cdot\! \frac{g(\theta)}{\text{sum}(\text{abs}(g(\theta)))} +\! \\ 
			\qquad \qquad \qquad \qquad \quad \ \gamma_2\! \cdot\! \frac{2}{\pi}\! \cdot\!  \text{arctan}(\frac{g(\theta)}{\text{mean}(\text{abs}(g(\theta)))}) +\! \\ 
			\qquad \qquad \qquad \qquad \quad \ \text{sign}(g(\theta)) \big) \Big ),
		\end{array}
		\right. 
	\label{eq:inner-loop}
\end{equation} 
where both $\gamma_1$ and $\gamma_2$ are set to 0.01 by default.
$\epsilon_c$ determines the perturbation scale.
$\mathcal{L}(\mathcal{U}_{\theta}(x_{\mathcal{D}_k}))$ is the entropy of $\mathcal{U}_\theta(x_{\mathcal{D}_k})$, and $g(\theta)$ is the gradient of the entropy \emph{w.r.t} $x_{\mathcal{D}_k}$ based on the current parameter $\theta$. 
Clip is the function that clips each pixel value of the image into the range of $[0, 255]$.

\begin{algorithm}[t]
	\caption{Training of Image Classification Eraser}
	\label{algorithm:ICE_train}
	{\bfseries input:} Source datasets $\mathbf{D} \!=\! \{\mathcal{D}_1, \mathcal{D}_2, ...,  \mathcal{D}_m\}$, Source models $\mathbf{M}_{\mathcal{D}_k} \!=\! \{ \mathbf{M}_{\mathcal{D}_k}^1, \mathbf{M}_{\mathcal{D}_k}^2, ..., \mathbf{M}_{\mathcal{D}_k}^{N_k} \}$ for each dataset $\mathcal{D}_k$. \\
	{\bfseries output:} Optimized weight $\theta$. \\
	{\bfseries 1 $\;\!$:} {\bfseries while not} done {\bfseries do} \\
	{\bfseries 2 $\;\!$:} 	\quad	\textbf{for} each $\mathcal{D}_k \in \mathbb{D}$ \textbf{do}\\
	{\bfseries 3 $\;\!$:} 	\qquad	Sample a mini data batch $(X_{\mathcal{D}_k}, Y_{\mathcal{D}_k}) \in \mathcal{D}_k$ \\
	{\bfseries 4 $\;\!$:} 	\qquad	Obtain adversarial examples $\hat{X}_{\mathcal{D}_k}$ via Eq.~\ref{eq:inner-loop} \\
	{\bfseries 5 $\;\!$:}     \quad\quad \textbf{for} each $\mathbf{M}_{\mathcal{D}_k}^j \in \mathbf{M}_{\mathcal{D}_k}$ \textbf{do} \\
	{\bfseries 6 $\;\!$:}     \qquad\quad Obtain adversarial loss $\mathbf{L}_{\mathcal{D}_k}^{j}$ for $\hat{X}_{\mathcal{D}_k}$ via Eq.~\ref{eq:evaluation}. \\
	{\bfseries 7 $\;\!$:}     \qquad \textbf{end for} \\
	{\bfseries 8 $\;\!$:}     \quad \textbf{end for} \\
	{\bfseries 9 $\;\!$:}  \quad   $\theta \!=\! \theta \!+\! \alpha\!\cdot\!\nabla_{\theta} \big( \frac{1}{m}\sum_{k=1}^{m} (\frac{1}{N_k} \sum_{j=1}^{N_k} \mathbf{L}_{\mathcal{D}_k}^{j}) \big)$ \\
	{\bfseries 10:} 	{\bfseries end while} \\
	{\bfseries 11:} 	{\bfseries return} $\theta$
\end{algorithm}

\textbf{In the outer-Loop}, 
we evaluate the attack success rate of the adversarial examples generated under the GTA setting and optimize the success rate by optimizing the model $\mathcal{U}_\theta$.
Specifically, we evaluate how the perturbed image $\hat{x}_{\mathcal{D}_k}$ generated in inner-loop fools each simulated unknown target model $\mathbf{M}_{\mathcal{D}_k}^j \in \mathbf{M}_{\mathcal{D}_k}$ by calculating the adversarial loss 
\begin{equation}
	\!\!\!\!\!\! \bm{l}_{\mathcal{D}_k}^{j} \!\!\!\!=\! \text{CE}(\mathbf{M}_{\mathcal{D}_k}^j \!\! (\hat{x}_{\mathcal{D}_k}\!), y_{\mathcal{D}_k}\!),
	\label{eq:evaluation}
\end{equation}
where $j \in [1, N_k]$;
$y_{\mathcal{D}_k}$ is the groundtruth of $x_{\mathcal{D}_k}$; CE is cross-entropy;  $\mathbf{M}_{\mathcal{D}_k}^j(\hat{x}_{\mathcal{D}_k})$ is the target model's prediction for $\hat{x}_{\mathcal{D}_k}$.
A larger adversarial loss $\bm{l}_{\mathcal{D}_k}^j$  indicates a higher possibility that the simulated unknown target model $\mathbf{M}_{\mathcal{D}_k}^j$ is fooled by $\hat{x}_{\mathcal{D}_k}$.
Note that the groundtruths of all simulated unknown images are accessible when we training the model $\mathcal{U}_\theta$, so we use cross-entropy instead of entropy used in inner-loop to calculate the loss in outer-loop.

To ensure that the classification information of each image $x_{D_k}\in \mathcal{D}_k$ can be erased by attacking model $\mathcal{U}_\theta$ and to ensure that the perturbed image $\hat{x}_{D_k}$ is confusing for the simulated unknown target model $\mathbf{M}_{\mathcal{D}_k}^j$ to predict, we optimize the model $\mathcal{U}_\theta$ by maximizing the adversarial loss $\bm{l}_{\mathcal{D}_k}^j$ by the following SGD update:
\begin{equation}
	\theta = \theta + \alpha \cdot
	\nabla_{\theta} \bm{l}_{\mathcal{D}_k}^j,
\end{equation}
where $\alpha$ is the learning rate.
$\bm{l}_{\mathcal{D}_k}^j$ is differentiable \emph{w.r.t} $\theta$ because $\bm{l}_{\mathcal{D}_k}^j$ depends on $\hat{x}_{\mathcal{D}_k}$ and $\hat{x}_{\mathcal{D}_k}$ depends on $\theta$.
In our experiments, we optimize the model $\mathcal{U}_\theta$ by simultaneously maximizing the adversarial losses on all source models in each iteration, which is summarized in Algorithm~\ref{algorithm:ICE_train}.
This procedure will enforce $\mathcal{U}_\theta$ having the property that for clean images from unknown datasets, the corresponding adversarial examples constructed by attacking $\mathcal{U}_\theta$ are effective to fool unknown target models.

\subsection{Inference and evaluation}
\label{sec:inference of ICE}
Given any clean unknown image $x$ that will be fed into an unknown target model $\mathbb{M}$, we evaluate the proposed ICE with the following steps.
\textbf{1)} Directly feed the image $x$ into model $\mathcal{U}_\theta$ and generate the adversarial example $x^{(T)}$ by maximizing the entropy for $T$ gradient ascent steps.
The $i$-th step is formulated as
\begin{equation}
	\left\{
	\begin{array}{lr}
		\delta^{(i-1)} = \text{sign}\big (\nabla_{x^{(i-1)}}\mathcal{L}(\mathcal{U}_\theta(x^{(i-1)}) ), \\
		x^{(i)} = \text{clip}(x^{(i-1)} + \frac{\epsilon}{T} \cdot \delta^{(i-1)}), 
	\end{array}
	\right.
	\label{eq:generation of the perturbation}
\end{equation}
where $x^{(0)}=x$ and $\delta^{(i-1)}$ is the perturbation generated in the $i$-th step.
$\epsilon/T$ is the $L_\infty$ perturbation scale in each step.
\textbf{2)} Generate the adversarial example $\hat{x}$ by the formulation 
\begin{equation}
	\hat{x} = \text{clip}(x+\epsilon \cdot \text{sign}(x^{(T)} - x))
	\label{eq:sign projection}
\end{equation}
We call this step `Sign-Projection' (SP).
The reason why we use SP is that it enlarges the average distortion of each pixel without amplifying the $L_\infty$ norm of the perturbation, which improves the GTA success rate.
Ablation study in Section~\ref{sec:adlation-SP} will show that SP improves both ICE and the baselines introduced in Section~\ref{sec:evaluate of baselines}. 
\textbf{3)} Feed the adversarial example $\hat{x}$ and the clean image $x$ into the unknown target model $\mathbb{M}$ and get its predictions $\mathbb{M}(\hat{x})$ and $\mathbb{M}(x)$.
\textbf{4)} The GTA process is successful if $\mathbb{M}(\hat{x}) \neq \mathbb{M}(x)$.
The evaluation pipeline is also illustrated in Figure~\ref{fig:framework}b, where the green arrows indicate the second step.

\section{Sine Attack}
\label{sec:method-SA}
Experiments in Section~\ref{sec:experiment} will demonstrate that ICE significantly outperforms the baselines under the GTA setting.
Despite the clearly improved GTA performance achieved by ICE, another great interesting phenomenon is that compared with the other methods, ICE generates similar texture-like noises for different images, even when the images are sampled from different categories and with different resolutions, as Fig.\ref{fig:fig1} and Section~\ref{sec:visualization} show.
To further investigate ICE noise, we visualize some ICE noises in frequency domain via FFT.
Fig~\ref{fig:fig1} shows some spectrum diagram examples of ICE noise and Section~\ref{sec:visualization} will show more spectrums.
Obviously, each ICE noise contains several sine signals and the most important components are three sine waves with certain frequencies for the three R, G, B image channels, respectively.

\begin{figure}[t]
	\centering
	\includegraphics[width=0.5\linewidth]{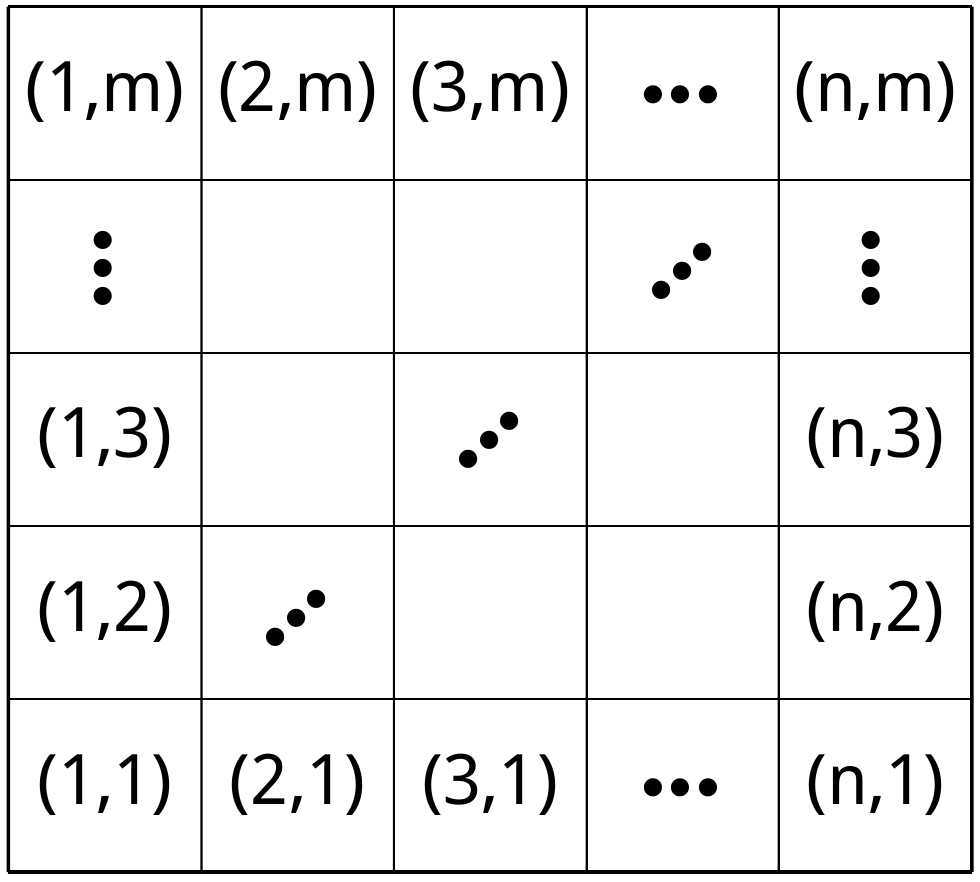}
	\caption{
		The ($X_{map}$, $Y_{map}$) coordination map with resolution $n\times m$.
		Each pixel's coordination is denoted as ($x_{map}$,$y_{map}$).
	}
	\label{fig:X_Y}
\end{figure}

Inspired by this interesting finding, we design another simple yet effective method called Sine Attack (SA) which uses the three sine signals to build an attacker and optimizes the attacker's fooling rate under the GTA setting by optimizing the three sine waves.
We formulate each sine wave as
\begin{equation}
	Z_j=Sine(a_j\cdot X_{map}+b_j\cdot Y_{map}+c_j),
	\label{eq:sa1}
\end{equation}
where $j\in$ [R, G, B].
($X_{map}$, $Y_{map}$) denote the coordinate of all pixels, as Fig.~\ref{fig:X_Y} shows. 
$a_j, b_j, c_j$ are three scalars of the sine wave in $j$-channel.
They together determine the sine wave frequency, direction, and initial phase.
That is to say, SA optimizes the three sine waves by optimizing 
the nine scalars: $a_R$, $b_R$, $c_R$, $a_G$, $b_G$, $c_G$, $a_B$, $b_B$, $c_B$.
For simplicity, we use $\bm{\omega}$ to represent the vector of the nine scalars in the rest of the paper and formulate the attacker constructed by the three sine waves as
\begin{equation}
	\mathcal{Z}(\bm{\omega}, X_{map}, Y_{map}) = \text{Concat}(Z_R, Z_G, Z_B).
	\label{eq:SA-concat}
\end{equation} 

\subsection{Training of SA}
We also optimize $\bm{\omega}$ on the resources introduced in Section~\ref{sec:GTA}: source datasets $\mathbf{D} \!=\! \{\mathcal{D}_1, \mathcal{D}_2, ...,  \mathcal{D}_m\}$ and source models $\mathbf{M}_{\mathcal{D}_k} \!=\! \{ \mathbf{M}_{\mathcal{D}_k}^1, \mathbf{M}_{\mathcal{D}_k}^2, ..., \mathbf{M}_{\mathcal{D}_k}^{N_k} \}$ for each source dataset $\mathcal{D}_k$.
Given any source dataset $D_k$ and any image $x_{\mathcal{D}_k} \in \mathcal{D}_k$, we simulate it as a randomly encountered image and use the following three steps to optimize $\bm{\omega}$.

\textbf{First}, we create the coordinate map ($X_{map}$, $Y_{map}$) for $x_{\mathcal{D}_k}$ and use the noise $\mathcal{Z}(\bm{\omega}, X_{map}, Y_{map})$ to conduct a GTA process on $x_{\mathcal{D}_k}$ to obtain the resulting adversarial image.
We formulate this process as 
\begin{equation}
	\hat{x}_{\mathcal{D}_k} \!\! = \! \text{Clip}(x_{\mathcal{D}_k} \!\!+ \epsilon \cdot \mathcal{Z}(\bm{\omega}, X_{map}, Y_{map}) ).
	\label{eq:SA-train-attack}
\end{equation}

Compared with ICE, the creation of adversarial images with SA is much simpler than that in ICE because SA crafts the adversarial image by directly adding the perturbation to the victim image without calculating gradient.

\textbf{Second}, we simulate each source model $\mathbf{M}_{\mathcal{D}_k}^j \in \mathbf{M}_{\mathcal{D}_k}$ as an unknown victim model and evaluate how the perturbed image $\hat{x}_{\mathcal{D}_k}$ fools each simulated unknown victim model by calculating the adversarial loss 
\begin{equation}
	\!\!\!\!\!\! \bm{l}_{\mathcal{D}_k}^{j} \!\!\!\!=\! \text{CE}(\mathbf{M}_{\mathcal{D}_k}^j \!\! (\hat{x}_{\mathcal{D}_k}\!), y_{\mathcal{D}_k}\!).
	\label{eq:SA-train-evaluate}
\end{equation}

\textbf{Third}, the goal of SA is that the noise $\mathcal{Z}(\bm{\omega}, X_{map}, Y_{map})$ can effectively perturb the victim image so that the victim model can be fooled.
Therefore we optimize the attacker by maximizing the above adversarial loss, which can be formulated as
\begin{equation}
	\begin{aligned}
		\bm{\omega} \!&=\! \bm{\omega} \!+\! \alpha \!\cdot\!\nabla_{\bm{\omega}} \bm{l}_{\mathcal{D}_k}^j \\
		\!&=\! \bm{\omega} \!+\! \alpha \!\cdot\!\nabla_{\bm{\omega}} \text{CE}(\mathbf{M}_{\mathcal{D}_k}^j \!\! (\hat{x}_{\mathcal{D}_k}\!), y_{\mathcal{D}_k}\!) \\
		\!&=\! \bm{\omega} \!+\! \alpha \!\cdot\!\nabla_{\bm{\omega}}\hat{x}_{\mathcal{D}_k} \!\cdot\! \nabla_{\hat{x}_{\mathcal{D}_k}}\text{CE}(\mathbf{M}_{\mathcal{D}_k}^j \!\! (\hat{x}_{\mathcal{D}_k}\!), y_{\mathcal{D}_k}\!) \\
		\!&=\! \bm{\omega} \!+\! \alpha \!\cdot\!\nabla_{\bm{\omega}}(\text{Clip}(x_{\mathcal{D}_k} \!\!+ \epsilon \!\cdot\! \mathcal{Z}(\bm{\omega}, X_{map}, Y_{map})) \cdot \\
		&\qquad \qquad \qquad \qquad \qquad \qquad  \nabla_{\hat{x}_{\mathcal{D}_k}}\text{CE}(\mathbf{M}_{\mathcal{D}_k}^j \!\! (\hat{x}_{\mathcal{D}_k}\!), y_{\mathcal{D}_k}\!), 
	\end{aligned}
	\label{eq:SA-train-optimize}
\end{equation}
where
$\alpha$ is the learning rate.
Eq.\ref{eq:SA-train-optimize} shows that when we optimize $\bm{\omega}$, we need to calculate the gradient $\nabla_{\bm{\omega}}(\text{Clip}(x_{\mathcal{D}_k} \!\!+ \epsilon \!\cdot\! \mathcal{Z}(\bm{\omega}, X_{map}, Y_{map}))$, so we do not use sign function to discrete each pixel of $\mathcal{Z}(\bm{\omega}, X_{map}, Y_{map})$ to -1 or 1 in Eq.\ref{eq:SA-train-attack}. 
Note that in our experiments, we optimize $\bm{\omega}$ by simultaneously maximizing the adversarial losses on all source models in each iteration, which is summarized in Algorithm~\ref{algorithm:SA_train}.

\begin{algorithm}[t]
	\caption{Training of Sine Attacker}
	\label{algorithm:SA_train}
	{\bfseries input:} Source datasets $\mathbb{D} \!=\! \{\mathcal{D}_1, \mathcal{D}_2, ...,  \mathcal{D}_m\}$, Source models $\mathbf{M}_{\mathcal{D}_k} \!=\! \{ \mathbf{M}_{\mathcal{D}_k}^1, \mathbf{M}_{\mathcal{D}_k}^2, ..., \mathbf{M}_{\mathcal{D}_k}^{N_k} \}$ for each dataset $\mathcal{D}_k$. \\
	{\bfseries output:} Optimized perturbation parameter $\bm{\omega}$. \\
	{\bfseries 1 $\;\!$:} {\bfseries while not} done {\bfseries do} \\
	{\bfseries 2 $\;\!$:} 	\quad	\textbf{for} each $\mathcal{D}_k \in \mathbb{D}$ \textbf{do}\\
	{\bfseries 3 $\;\!$:} 	\qquad	Sample a mini data batch $(X_{\mathcal{D}_k}, Y_{\mathcal{D}_k}) \in \mathcal{D}_k$ \\
	{\bfseries 4 $\;\!$:}   \qquad  Create ($X_{map}$, $Y_{map}$) for all image $\in$ $X_{\mathcal{D}_k}$ \\
	{\bfseries 5 $\;\!$:} 	\qquad	Obtain adversarial examples $\hat{X}_{\mathcal{D}_k}$ via Eq.~\ref{eq:SA-train-attack} \\
	{\bfseries 6 $\;\!$:}     \quad\quad \textbf{for} each $\mathbf{M}_{\mathcal{D}_k}^j \in \mathbf{M}_{\mathcal{D}_k}$ \textbf{do} \\
	{\bfseries 7 $\;\!$:}     \qquad\quad Obtain adversarial loss $\mathbf{L}_{\mathcal{D}_k}^{j}$ on $\hat{X}_{\mathcal{D}_k}$ via Eq.~\ref{eq:SA-train-evaluate}. \\
	{\bfseries 8 $\;\!$:}     \qquad \textbf{end for} \\
	{\bfseries 9 $\;\!$:}     \quad \textbf{end for} \\
	{\bfseries 10:}  \quad   $\bm{\omega} \!=\! \bm{\omega} \!+\! \alpha\!\cdot\!\nabla_{\bm{\omega}} \big( \frac{1}{m}\sum_{k=1}^{m} (\frac{1}{N_k} \sum_{j=1}^{N_k} \mathbf{L}_{\mathcal{D}_k}^{j}) \big)$ \\
	{\bfseries 11:} 	{\bfseries end while} \\
	{\bfseries 12:} 	{\bfseries return} ${\bm{\omega}}$
\end{algorithm}

\subsection{Inference and evaluation}
\textbf{Note that SA is a natural universal adversarial perturbation that use the optimized three sine waves to perturb all possible images.}
Therefore, compared with ICE which generates image-dependent perturbations, SA perturbs victim images simpler in inference. 
Given any clean image $x$ that will be fed into an unknown target model $\mathbb{M}$, we evaluate the proposed SA with the following steps.
\textbf{1)} Create the coordination map ($X_{map}$, $Y_{map}$) for $x$.
\textbf{2)} Use the trained $\bm{\omega}$ to generate three sine waves and construct the noise $\mathcal{Z}(\bm{\omega}, X_{map}, Y_{map})$ via Eq.\ref{eq:SA-concat}.
\textbf{3)} 
Generate the adversarial example $\hat{x}$ by the formulation 
\begin{equation}
	\hat{x} = \text{clip}\Big(x+\epsilon \cdot \text{sign}\big(\mathcal{Z}(\bm{\omega}, X_{map}, Y_{map})\big)\Big)
	\label{eq:SA_sign projection}
\end{equation}
\textbf{4)} Feed the adversarial example $\hat{x}$ and the clean image $x$ into the unknown target model $\mathbb{M}$ and get its predictions $\mathbb{M}(\hat{x})$ and $\mathbb{M}(x)$.
\textbf{5)} The GTA process is successful if $\mathbb{M}(\hat{x}) \neq \mathbb{M}(x)$.

\section{Experimental Results}
\label{sec:experiment}
In this section, we conduct several experiments to evaluate the proposed methods for conducting generalized transferable attacks. 
Four datasets Cifar-10~\cite{krizhevsky2009learning}, Cifar-100, Tiered$_{V56}$, and Tiered$_{T84}$ are used to build the basic GTA testing scenes.
More challenging GTA testing scenes on other datasets (\emph{e.g.}, ImageNet~\cite{deng2009imagenet}, CUB~\cite{WahCUB_200_2011}, MS-COCO\cite{lin2014microsoft}) will be introduced in Sections~\ref{sec:attacking CUB}, \ref{sec:attacking ImageNet}, and \ref{sec:attacking object detection}.
Both Cifar-10 and Cifar-100 contain 60,000 images with the resolution of 32$\times$32.
Tiered$_{V56}$ and Tiered$_{T84}$ are two sub-datasets of TieredImageNet~\cite{ren2018meta}.
TieredImageNet is a subset sampled from ImageNet~\cite{deng2009imagenet} and contains 351, 97, and 160 image classes in training, validation, and testing, respectively.
We split TieredImageNet into Tiered$_{V56}$ and Tiered$_{T84}$.
Tiered$_{T84}$ contains all images from the training set of TieredImageNet and all the images are resized into 84$\times$84 resolution.
Tiered$_{V56}$ contains all images from the validation and test sets of TieredImageNet and all the images are resized into 56$\times$56 resolution.
There is no overlapping image category between Tiered$_{V56}$ and Tiered$_{T84}$.
For either Tiered$_{T84}$ or Tiered$_{V56}$, we 
use the first 1200 images of each category to compose the training set and use the last 100 images to compose the testing set. 
More details of {Tiered$_{T84}$}, and {Tiered$_{V56}$} will be shown in Appendix.

\begin{figure}[t]
	\centering
	\includegraphics[width=0.95\linewidth]{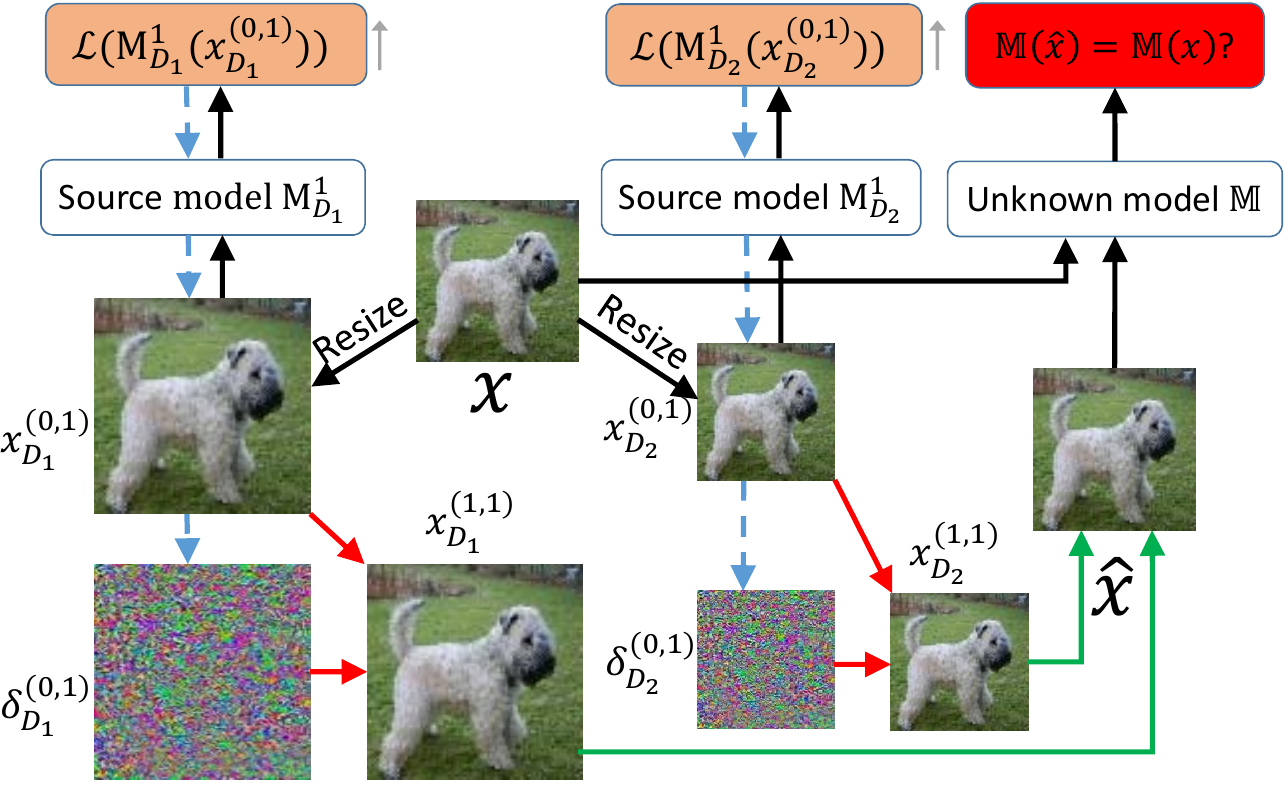}
	\caption{
		Inference pipeline of PGD-based baselines.
	}
	\label{fig:test_pipeline}
\end{figure}

\subsection{Baselines}
\label{sec:baselines}
GTA is a novel adversarial attack problem and few existing methods can be directly used as baselines.
Considering that transfer attack is a similar problem to GTA, we deploy several transferable adversarial attack methods including MI~\cite{dong2018boosting}, DI~\cite{xie2019improving}, TI-DIM~\cite{dong2019evading}, AEG~\cite{bose2020adversarial},  IR~\cite{wang2021unified}, MTA~\cite{qin2021training}, and FDA+xent\cite{nathan2020perturbing} on GTA as baselines for the proposed ICE and SA.
Despite the transfer attack methods, we also consider some other methods as baselines, such as FGSM, PGD, UAP\cite{zhang2021data}, and the cross domain adversarial attack method RAP\cite{naseer2019cross-domain}.

\subsubsection{How to use baselines for GTA?}
\label{sec:evaluate of baselines}
The baselines FGSM, PGD, MI, DI, and TI-DIM are default implemented on all our GTA experiments (including the adversarial attack to MS-COCO in Section~\ref{sec:attacking object detection}) with the inference pipeline described below.
More kinds of implementation pipelines (\emph{e.g.}, using KL-divergence, optimizing a single adversarial image) will be shown in Appendix.
The detailed implementations of AEG, IR, MTA, FDA+xent, UAP, and RAP will also be introduced in Appendix.

1) Since the source models trained on different source datasets commonly have different input shapes, we firstly resize the victim image $x$ to the input shapes of all source models and then feed the resized images to the source models, respectively. 
Then the inference of $x$ on each source model $\mathbf{M}_{D_k}^j \in \mathbf{M}_{D_k}$ can be formulated as	$y_{D_k}^{(0, j)} = \mathbf{M}_{D_k}^j(x_{D_k}^{(0,j)})$, where $x_{D_k}^{(0,j)} = \text{resize}(x, \text{resolution}(\mathbf{M}_{D_k}^j))$.
$\text{resolution}(\mathbf{M}_{D_k}^j)$ means the resolution of $\mathbf{M}_{D_k}^j$.
2) Because we cannot access the category of the image $x$ in advance, no ground-truth label can be leveraged to perturb the resized image.
Therefore, for each source model, we generate adversarial perturbation by maximizing the entropy (as used in our ICE) for $T$ gradient ascent steps.
The $i$-th step is
\begin{equation}
	\left\{
	\begin{array}{lr}
		y_{D_k}^{(i-1,j)} = \mathbf{M}_{D_k}^j(x_{D_k}^{(i-1,j)}), \\
		\delta_{D_k}^{(i-1,j)} = \text{sign}\big (\nabla_{x_{D_k}^{(i-1,j)}}\mathcal{L}(y_{D_k}^{(i-1,j)}) ), \\
		x_{D_k}^{(i,j)} = \text{clip}(x_{D_k}^{(i-1,j)} + \frac{\epsilon}{T} \cdot \delta_{D_k}^{(i-1,j)}),
	\end{array}
	\right. 
	\label{eq:generation of the perturbation for baseline}
\end{equation}
where $y_{D_k}^{(i-1,j)}$ is the source model's output and $x_{D_k}^{(i,j)}$ and
$\delta_{D_k}^{(i-1,j)}$ are the adversarial example and the perturbation generated in the $i$-th step, respectively.
3) We resize all the adversarial examples to the image $x$'s original shape and average fuse them to a single image $x_{adv}$ following the formulation $x_{adv} = \frac{1}{m} \cdot \sum_{k=1}^{m} \cdot \big ( \frac{1}{N_k} \cdot  \sum_{j=1}^{N_k}\text{resize}(x_{D_k}^{(T,j)}, \text{resolution}(x)) \big)$.
4) We generate adversarial example $\hat{x}$ by the formulation $\hat{x} = \text{clip}(x+\epsilon \cdot \text{sign}(x_{adv} - x))$, which is the SP step defined in Section~\ref{sec:inference of ICE}. 
5) We feed the adversarial example $\hat{x}$ and the clean image $x$ into the unknown target model $\mathbb{M}$ and get its predictions.
6) The GTA process is successful if $\mathbb{M}(\hat{x}) \neq \mathbb{M}(x)$.
The evaluation of the PGD-based baselines is also illustrated in Fig.~\ref{fig:test_pipeline}, where green arrows denote the third and fourth steps.

\subsection{Experimental settings}
\label{sec:experimental settings}
1) {\bf Source and Target models}. 
On each dataset, we train several models including VGG-16~\cite{simonyan2015very}, ResNet-18~\cite{he2016deep}, MobileNet-V3~\cite{howard2019searching}, and DenseNet-26~\cite{huang2017densely}.
The last three models will be simplified as RN-18, MB-V3, and DN-26, respectively, in the rest of our paper.
All these models will be used as the source and target models in GTA testing experiments.
The training details and the architectures of these models will be introduced in Appendix.

2) {\bf Hyper-parameters in the training phase of ICE}.
In the training phase of ICE, we train $\mathcal{U}_\theta$ for 50,000 iterations with batch size of 64 for each source dataset.
Learning rate $\alpha$ is set to 0.01.
$\epsilon_c$ is set to 3000 and is periodically decayed by 0.9$\times$ for every 3000 iterations.

3) {\bf Hyper-parameters in the training phase of SA}.
We train $\bm{\omega}$ for 10,000 iterations with batch size of 64 for each source dataset.
Learning rate $\alpha$ is set to 0.01.

4) {\bf Evaluation}.
In the inference phase, we set $T$ to 10 by default for ICE and other PGD-based baselines, and set $\epsilon$ to 15 for all methods.
Section~\ref{sec:smaller_eps} will show the experimental results with $\epsilon=8$, where ICE and SA still outperform the baselines with significant margins.
Note that we use \textbf{attack success rate} to denote the GTA performance by default.
Because attacking the images that are wrongly classified by the target model is meaningless, we only attack the images that are correctly classified by the target model.

\subsection{Generalized transferable attack to Cifar-100}
\label{sec:attack cifar100}
In this subsection, we perform GTA experiments on Cifar-100, i.e., perturbing Cifar-100 images, to fool the target victim models. 
The models MB-V3, VGG-16, RN-18, and DN-26 trained on Cifar-100 are used as victims to calculate the GTA success rate.
Different settings of source models and datasets are evaluated and experimental results are reported in Table \ref{tab:attack on cifar100}.

In the first experiment (the first row of Table~\ref{tab:attack on cifar100}), we use RN-18 and Cifar-10 as the source model and the source dataset, respectively, and conduct GTA on the testing images from Cifar-100.
The proposed ICE and SA outperform all existing methods on the GTA problem.
Compared with UAP, ICE and SA promote the average attack success rate on the four target models by about 1.1\% and 16.6\%, respectively.
Moreover, SA outperforms ICE with clear margins, indicates that the optimized three sine waves are enough to perturb Cifar-100 images.
We guess the reasons why SA outperforms ICE are \textbf{1)} GTA is a much challenging setting where the attacker must be generalized enough to perturb any unknown images; \textbf{2)} the optimized three sine waves in SA are the most crucial noises to solve GTA; \textbf{3)} despite the three sine waves, ICE uses more complex noises to solve GTA which may improve ICE's performance in training phase but damage the generalization.
In other words, \textbf{compared with ICE, SA which uses only three sine waves is more generalized to perturb unknown images.}

\begin{table}[t!]
	\centering
	\caption{GTA success rates on Cifar-100.
	}
	\setlength\tabcolsep{5pt}
	\scriptsize   
	\begin{tabular}{cccccc}
		\toprule
		Resource & Method & MB-V3 & VGG-16 & RN-18 & DN-26\\
		\midrule
		\multirow{7}{*}{ \tabincell{c}{Cifar-10 \\ (Res-18)} }
		&FGSM 		      &47.7\%     &64.2\% & 59.1\%   & 73.1\% \\
		&PGD 			   &37.3\%    &50.9\%  & 43.2\%  & 61.5\%\\
		&DI 			   &39.3\%    &54.9\%  & 47.2\%  & 64.7\%\\
		&MI 			   &45.4\%    &62.3\%  & 56.3\%  & 72.1\%\\
		&TI-DIM 			   &51.0\%    &45.6\%  & 48.0\%  & 55.3\%\\
		&IR         &48.3\%    &62.1\%  & 60.5\% & 73.7\%\\
		&AEG  	     &51.3\%    &61.7\%  & 61.2\%  & 66.4\%\\
		&MTA  	     &42.3\%    &49.5\%  & 45.0\%  & 60.0\% \\
		&UAP  	     &53.9\%    &69.7\%  & 71.2\%  & 79.5\% \\
		&RAP  	     &50.5\%    &68.8\%  & 70.7\%  & 81.3\% \\
		&FDA+xent  	     &37.8\%    &51.5\%  & 44.3\%  & 62.6\% \\
		\rowcolor {gray!20} \cellcolor{gray!0} &\textbf{ICE}    &{54.9\%}    &{67.3\%}  & {71.5\%}  & {83.5\%}\\
		\rowcolor {gray!20} \cellcolor{gray!0} &\textbf{SA}    &\textbf{78.3\%}    &\textbf{77.8\%}  & \textbf{79.6\%}  & \textbf{84.2\%}\\
		\midrule
		\multirow{6}{*}{\tabincell{c}{Cifar-10 \\ + Tiered$_{T84}$ \\ (Res-18)}}
		&FGSM 		          &33.0\%     &47.3\% & 38.6\%  & 58.7\% \\
		&PGD 		  &39.0\%    &52.2\%  & 41.7\%  & 64.9\% \\
		&DI  	  &39.1\%    &52.3\%  & 41.7\%  & 64.2\% \\
		&MI  		 &42.7\%    &57.0\%  & 48.2\%  &68.1\%\\
		&TI-DIM  	  &55.1\%    &48.2\%  & 50.5\%  & 58.1\% \\
		&IR        &44.0\%    &58.1\%  & 51.9\%   & 69.5\%\\
		&AEG      	  &50.9\%    &64.6\%  & 58.2\%  & 70.3\%\\
		&MTA  	 &43.6\%    &56.7\%  & 47.5\%  & 68.0\% \\
		&UAP  	     &51.1\%    &73.7\%  & 74.2\%  & 81.5\% \\
		&RAP  	     &46.2\%    &70.3\%  & 67.0\%  & 78.6\% \\
		&FDA+xent  	     &38.8\%    &52.0\%  & 42.6\%  & 65.2\% \\
		\rowcolor {gray!20} \cellcolor{gray!0}&\textbf{ICE} &{57.0\%}    &{76.2\%}  & {77.5\%}  & {83.3\%} \\
		\rowcolor {gray!20} \cellcolor{gray!0} &\textbf{SA}    &\textbf{68.8\%}    &\textbf{78.2\%}  & \textbf{80.7\%}  & \textbf{85.2\%}\\
		\midrule
		\multirow{6}{*}{\tabincell{c}{Cifar-10 \\ + Tiered$_{V56}$ \\ (RN-18)}}
		&FGSM 		          &33.7\%     &48.5\% & 38.9\%  &58.2\% \\
		&PGD 		  &40.1\%    &54.9\%  & 44.3\%  & 65.7\%\\
		&DI  	  &40.5\%    &55.0\%  & 44.7\%  & 65.9\% \\
		&MI  		 &45.5\%    &60.3\%  & 50.2\%  & 70.6\%\\
		&TI-DIM  	  &51.4\%    &48.0\%  & 49.4\%  & 58.9\% \\
		&IR        &44.7\%    &59.2\%  & 54.0\% & 70.1\%\\
		&AEG      	  &51.5\%    &62.3\%  & 58.9\%  & 68.3\% \\
		&MTA  	 &43.3\%    &60.8\%  & 50.0\%  & 70.2\% \\
		&UAP  	     &51.2\%    &71.8\%  & 69.7\%  & 80.6\% \\
		&RAP  	     &46.5\%    &69.7\%  & 67.5\%  & 77.2\% \\
		&FDA+xent  	     &39.6\%    &52.3\%  & 42.9\%  & 65.6\% \\
		\rowcolor {gray!20} \cellcolor{gray!0}&\textbf{ICE} &{52.7\%}    &{76.9\%}  & {79.8\%} & {84.9\%}\\
		\rowcolor {gray!20} \cellcolor{gray!0} &\textbf{SA}    &\textbf{65.9\%}    &\textbf{77.3\%}  & \textbf{82.6\%}  & \textbf{86.0\%}\\
		\midrule
		\multirow{6}{*}{\tabincell{c}{Cifar-10 \\ + Tiered$_{T84}$ \\ + Tiered$_{V56}$  \\  (RN-18)}}
		&FGSM 		      &45.3\%     &59.6\% & 49.9\%  & 70.0\% \\
		&PGD 			   &40.6\%    &54.7\%  & 42.8\%  & 65.8\% \\
		&DI  	       &40.6\%    &54.9\%  & 42.9\%  & 66.0\% \\
		&MI  		  &45.0\%    &58.9\%  & 48.1\%  & 69.2\%\\
		&TI-DIM  	       &54.3\%    &48.7\%  & 50.1\%  & 58.7\% \\
		&IR         &48.9\%    &57.5\%  & 51.2\%  & 70.1\%\\
		&AEG         &47.7\%    &63.8\%  & 56.0\%  & 71.7\% \\
		&MTA  	     &40.6\%    &61.3\%  & 49.2\%  & 69.7\% \\
		&UAP  	     &46.9\%    &65.6\%  & 66.2\%  & 77.3\% \\
		&RAP  	     &47.2\%    &62.7\%  & 56.1\%  & 70.8\% \\
		&FDA+xent  	     &43.3\%    &53.8\%  & 45.0\%  & 66.2\% \\
		\rowcolor {gray!20} \cellcolor{gray!0}&\textbf{ICE}    &{56.3\%}    &\textbf{{83.2\%}}  & \textbf{{90.1\%}}  & \textbf{{92.4\%}} \\
		\rowcolor {gray!20} \cellcolor{gray!0}&\textbf{ICE$_{A\_I}$}    &{54.1\%}    &{80.9\%}  & {87.9\%}  & {91.7\%} \\
		\rowcolor {gray!20} \cellcolor{gray!0}&\textbf{ICE$_{C\_\text{Tiered}_{T84}}$}    &{57.8\%}    &{78.3\%}  & {84.9\%}  & {90.1\%} \\
		\rowcolor {gray!20} \cellcolor{gray!0}&\textbf{ICE$_{C\_\text{Tiered}_{V56}}$}    &{64.2\%}    &{82.1\%}  & {87.2\%}  & {89.8\%} \\
		\rowcolor {gray!20} \cellcolor{gray!0} &\textbf{SA}    &\textbf{{66.9\%}}    &{80.5\%}  & {85.0\%}  & {86.9\%}\\
		\bottomrule
	\end{tabular}
	\label{tab:attack on cifar100}
\end{table}

In the second experiment, we 
use Cifar-10 and Tiered$_{T84}$ as the source datasets, and use two RN-18 models respectively trained on source datasets as the source models.
Interestingly, the baselines' performances in this experiment are commonly worse than their performances in the first experiment.
The possible reason for this result is that the resolution of the images from Tiered$_{T84}$ is 84$\times$84, which differs greatly from the resolution of the target dataset Cifar-100.
In comparison, ICE's performance in this experiment is much better than its performance in the first experiment, which indicates that ICE can efficiently make use of all the resources to improve its performance despite the difference among the source datasets.
Moreover, UAP performs the best among all baselines and SA still outperforms ICE.

In the third experiment, we use two RN-18 models respectively trained on Cifar-10 and Tiered$_{T56}$ as the source models.
It is observed that most of the baselines' performances in this experiment are slightly better than their performances in the third experiment.
For instance, compared with the performances in the third experiment, the performance of FGSM in this experiment is improved by about 1.7\%.
The possible reason for this result is that compared with the resolution of Tiered$_{T84}$, the resolution of Tiered$_{V56}$ is closer to the resolution of Cifar-100. 
Compared with UAP, the proposed ICE and SA promote the average attack success rate by about 7.7\% and 14.1\%, respectively.

In the fourth experiment, we use three RN-18 models respectively trained on Cifar-10, Tiered$_{T84}$ and Tiered$_{V56}$ as the source models.
Compared with the best baseline UAP, ICE and SA promotes the average attack success rate by about 25.8\% and 24.7\%.
The possible reason why ICE outperforms SA in this experiment are \textbf{1)} this experiment contains more source datasets than the three previous experiments, and \textbf{2)} \textbf{ICE is more efficient than SA to leverage additional source datasets to improve its generalization}.
The results of ICE$_{A\_I}$, ICE$_{C\_T84}$, and ICE$_{C\_V56}$ will be discussed in Section~\ref{sec:transfer noises}.

\begin{table}[t]
	\centering
	\caption{The GTA success rates on Cifar-10, Tiered$_{T84}$, and Tiered$_{V56}$.
	}
	\setlength\tabcolsep{5pt}
	\footnotesize  
	\begin{tabular}{cccccc}
		\toprule
		Resource & Method & MN-V3 & VGG-16 & RN-18 & DN-26\\
		\midrule
		\multirow{6}{*}{\tabincell{c}{-Cifar-10 \\ (RN-18)}}
		&FGSM 		      &21.3\%     &28.3\% & 33.0\%  & 47.9\% \\
		&MI  		  &20.6\%    &28.7\%  & 33.2\%  & 47.8\% \\
		&IR         &19.8\%    &28.3\%  & 32.3\%  & 47.9\% \\
		&AEG  	     &24.4\%    &38.5\%  & 41.2\%  & 52.1\% \\
		&UAP  	     &22.6\%    &37.9\%  & 49.2\%  & 54.9\% \\
		&RAP  	     &23.0\%    &38.2\%  & 40.9\%  & 52.2\% \\
		\rowcolor {gray!20} \cellcolor{gray!0} &\textbf{ICE}    &\textbf{36.3\%}    &{45.5\%}  & {61.0\%} & {65.5\%}\\
		\rowcolor {gray!20} \cellcolor{gray!0} &\textbf{SA}    &{23.2\%}    &\textbf{52.5\%}  & \textbf{68.8\%}  & \textbf{70.9\%}\\
		\midrule
		\multirow{6}{*}{\tabincell{c}{-Tiered$_{T84}$ \\ (RN-18)}}
		&FGSM 		      &56.5\%     &62.1\% & 83.5\%  & 88.7\% \\
		&MI  		  & \textbf{58.2\%}   &61.8\%  & 82.6\%  & 88.3\% \\
		&IR         &52.5\%    &56.8\%  & 79.3\%  & 87.5\% \\
		&AEG         &47.8\%    &45.6\%  & 56.0\%  & 64.5\% \\
		&UAP  	     &57.3\%    &63.3\%  & 82.8\%  & 88.0\% \\
		&RAP  	     &47.7\%    &49.6\%  & 58.3\%  & 64.6\% \\
		\rowcolor {gray!20} \cellcolor{gray!0} &\textbf{ICE}    &{52.3\%}    &{73.0\%}  & {90.8\%}  & {93.5\%}\\
		\rowcolor {gray!20} \cellcolor{gray!0} &\textbf{SA}    &\textbf{63.5\%}    &\textbf{87.2\%}  & \textbf{97.5\%}  & \textbf{97.7\%}\\
		\midrule
		\multirow{6}{*}{\tabincell{c}{-Tiered$_{V56}$ \\ (RN-18)}}
		&FGSM 		      &65.7\%     &69.8\% & 78.1\%  & 86.9\% \\
		&MI  		  &64.6\%    &68.0\%  & 76.2\%  & 86.5\% \\
		&IR  	     &59.7\%    &67.1\%  & 74.3\%  & 83.5\%\\
		&AEG  	     &57.0\%    &67.2\%  & 72.7\%  & 78.6\% \\
		&UAP  	     &70.9\%    &65.7\%  & 72.3\%  & 82.3\% \\
		&RAP  	     &57.1\%    &67.3\%  & 72.2\%  & 78.7\% \\
		\rowcolor {gray!20} \cellcolor{gray!0} &\textbf{ICE}    &{72.0\%}    &{83.5\%}  & {91.7\%}  & {93.3\%}\\
		\rowcolor {gray!20} \cellcolor{gray!0} &\textbf{ICE$_{A\_I}$}    &\textbf{79.7\%}    &\textbf{90.7\%}  & {95.3\%}  & {96.0\%} \\
		\rowcolor {gray!20} \cellcolor{gray!0} &\textbf{ICE$_{C\_T84}$}    &{69.8\%}    &{88.3\%}  & \textbf{96.5\%}  & \textbf{98.1\%} \\
		\rowcolor {gray!20} \cellcolor{gray!0} &\textbf{SA}    &{71.3\%}    &{77.0\%}  & {91.2\%}  & {92.5\%}\\
		\bottomrule
	\end{tabular}
	\label{tab:attack across}
\end{table}

\subsection{Generalized transferable attack to Cifar-10, Tiered$_{T84}$, and Tiered$_{V56}$}
\label{sec:exp_attacking_across_datasets}
Now we show ICE and SA still outperform baselines when using other datasets as target images. 
Table \ref{tab:attack across} reports the experimental results. 
There are four datasets in total (Cifar-10, Cifar-100, Tiered$_{T84}$, and Tiered$_{V56}$), and 
each row (denoted as -target) shows the experiment where we conduct GTA on the target dataset by using RN-18 trained on the other three datasets as source models.
For example, the `-Cifar10' row denotes the experiment where we use the other three datasets and the three respectively trained RN-18 models as resources to conduct GTA to Cifar-10 images.
Table~\ref{tab:attack across} does not show the `-Cifar100' row because the corresponding results have been shown in the last row of Table~\ref{tab:attack on cifar100}.
Note that to save space, Table \ref{tab:attack across} ignores some unimportant baselines used in Table \ref{tab:attack on cifar100}.
It is clear that given three datasets and the models trained on the three datasets, ICE and SA outperform the baselines to attack unknown images from other datasets.
The results of ICE$_{A\_I}$ and ICE$_{C\_T84}$ will be discussed in Section~\ref{sec:transfer noises}.

\subsection{Universal perturbation across images, datasets, and resolutions}
\label{sec:transfer noises}

We have experienced with the transfer-ability of ICE and SA under the standard GTA experiments.
Fig.\ref{fig:fig1} shows that ICE generates similar texture-like noises for different images, even when the images are sampled from different datasets with different resolutions.
Inspired by this phenomenon, here we perform an experiment to study whether the ICE noise generated for each image can lead to universal perturbation across images, datasets, and resolutions.


To evaluate whether the ICE noise generated for each image is universal within the target dataset, we randomly sample 1000 testing images from the target dataset, and transfer the ICE noise generated for each sampled image to other 999 sampled images.
We conduct this experiment on Cifar-100 and Tiered$_{V56}$ and denote this experiment as ICE$_{A\_I}$. 
Table~\ref{tab:attack on cifar100}'s fourth raw and Table~\ref{tab:attack across}'s third raw show the experimental results, where ICE$_{A\_I}$ performs comparable to ICE, demonstrating that ICE noise generated for each image is universal to perturb other images within the same dataset.
Note that for each image, we transfer its ICE noise to the other 999 images, so there are 1000$\times$999 attack results on each unknown target model, and the reported result on each model is the average of the 1000$\times$999 results.

We conduct cross-dataset experiments to evaluate whether the ICE noise generated for each image is universal to directly attack images from other datasets with different resolutions.
Let's use $\mathbb{D}_a$ and $\mathbb{D}_b$ to denote the two datasets in each cross-dataset experiment scene.
Because each original dataset in our paper contains at least 10,000 testing images, using all testing images to run the cross-dataset experiments is expensive.
So for each original dataset, we sample 1000 testing images to construct a sub-dataset.
Therefore we can obtain two sub-datasets $\mathcal{D}_a \in \mathbb{D}_a$ and $\mathcal{D}_b \in \mathbb{D}_b$.
In the cross-dataset experiment, we transfer the ICE noise of each image $x_a \in \mathcal{D}_a$ to each image $x_b \in \mathcal{D}_b$ with the formulation $\hat{x}_b = x_b + \epsilon\cdot\text{sign}(\text{crop}(\delta_a))$, where $\delta_a=\hat{x}_a-x_a$ is the ICE noise generated for image $x_a$.
We denote the ICE noise transferred to $x_b$ as ICE$_{C\_\mathbb{D}_a}$.
The experimental results shown in Table~\ref{tab:attack on cifar100}'s fourth raw and Table~\ref{tab:attack across}'s third raw demonstrate that ICE noise generated for each image is universal across datasets and resolutions.
Note that the image resolution of $\mathcal{D}_a$ should be no smaller than that of $\mathcal{D}_b$.
Either $\mathcal{D}_a$ or $\mathcal{D}_b$ contains 1000 sampled testing images, so there are 1000$\times$1000 cross-dataset testing results on each unknown target model, and the reported result on each model is the average of the 1000$\times$1000 results.

\begin{table}[t]
	\centering
	\caption{GTA results on CUB models. }
	\footnotesize 
	\begin{tabular}{ccccc}
		\toprule
		Resource & Method & VGG-16 & RN-18 & DN-26 \\
		\midrule
		\multirow{6}{*}{\tabincell{c}{Cifar-10  \\  (RN-18)}}
		&FGSM 		      & 14.4\% &22.7\%     &32.6\% \\
		&DI 			  & 31.8\%  &49.2\%    &72.5\% \\
		&MI 			  & 13.6\% &22.0\%    &30.7\% \\
		&TI-DIM 		  & 40.9\% &62.8\%    &73.4\% \\
		&MTA     &42.6\%    &51.3\% &62.3\% \\
		&UAP     &18.2\%    &31.0\% &47.5\%\\
		&RAP     &45.9\%    &60.5\% &72.6\%\\
		\rowcolor {gray!20} \cellcolor{gray!0} &\textbf{ICE}    &{47.9\%}  &{66.8\%}    &{87.3\%} \\
		\rowcolor {gray!20} \cellcolor{gray!0} &\textbf{SA}    &\textbf{50.3\%}    &\textbf{92.5\%} & \textbf{95.2\%}\\
		\bottomrule
	\end{tabular}
	\label{tab:attack CUB}
\end{table}

\subsection{Attacking Fine-Grained classification models}
\label{sec:attacking CUB}
We evaluate whether the proposed ICE and SA can be directly transferred to disturb the images from fine-grained classification datasets.
In this experiment, the source dataset and source model are Cifar-10 and RN-18 respectively, and the target dataset is CUB~\cite{WahCUB_200_2011}, a fine-grained bird classification dataset that differs greatly from Cifar-10.
The target models are VGG-16, RN-18, and DN-26 trained on CUB.
The three target models use a consistent input resolution of 112$\times$112 and achieve approximately 49.2\%, 54.9\%, and 48.3\% testing accuracies.
The attack success rates on the three target models are reported in Table~\ref{tab:attack CUB}.
Obviously, SA outperforms ICE, and ICE outperforms the other methods.

\subsection{Attacking ImageNet models}
\label{sec:attacking ImageNet}
Here we transfer the proposed ICE and SA to perturb victim images from the 392 ImageNet categories that do not belonging to TieredImageNet.
In this experiment, we use three RN-18 models respectively trained on Cifar-10, Tiered$_{T84}$ and Tiered$_{V56}$ as the source models, and use the publicly trained\footnote{https://github.com/tensorflow/models/tree/r1.12.0/research/slim} ResNet-101, DenseNet-161, Inception-V1, and Inception-V3 as the target models.
The four models will be shortened as RN-101, DN-161, Inc-V1, and Inc-V3, respectively.
Table~\ref{tab:attack ImageNet} reports the experimental results and demonstrates that the proposed ICE and SA still outperform the baselines in perturbing images that are sampled from novel categories and are using much larger resolutions (224$\times$224, 299$\times$299) than the source datasets.
\begin{table}[t]
	\centering
	\caption{GTA results on 392 ImageNet categories. }
	\footnotesize 
	\setlength\tabcolsep{5pt}
	\begin{tabular}{cccccc}
		\toprule
		Resource & Method  & RN-101 & DN-161 & Inc-V1 & Inc-V3 \\
		\midrule
		\multirow{8}{*}{\tabincell{c}{Cifar-10 \\ + Tiered$_{T84}$ \\ + Tiered$_{V56}$  \\  (RN-18)}}
		&FGSM  		 & 8.9\% & 6.6\% & 8.5\% & 6.7\% \\
		&DI  		 & 8.0\% & 6.5\% & 6.9\% & 5.8\% \\
		&MI  		 & 9.3\% & 7.6\% & 9.4\% & 7.9\% \\
		&TI-DIM  			& 21.9\% & 18.7\% & 22.0\% & 16.9\%\\
		&MTA 				& 19.6\% & 16.3\% & 22.5\% & 17.2\%\\
		&RAP  			& 23.9\% & 19.5\% & 23.0\% & 18.3\%\\
		&UAP				& 21.2\% & 15.8\% & 17.1\% & 17.8\%\\
		\rowcolor {gray!20} \cellcolor{gray!0} & \textbf{ICE}  	  &  \textbf{30.8\%} &  \textbf{30.5\%} & {35.0\%} & {38.3\%}\\
		\rowcolor {gray!20} \cellcolor{gray!0} & \textbf{SA}  	  &  {30.6\%} &  {23.8\%} & \textbf{{43.3\%}} & \textbf{{42.5\%}}\\
		\bottomrule
	\end{tabular}
	\label{tab:attack ImageNet}
\end{table}

\begin{table}[t]
	\centering
	\caption{Ablation GTA experiments on Cifar-100.
	}
	\footnotesize
	\begin{tabular}{cccccc}
		\toprule
		Setting & Method & MN-V3 & VGG-16 & RN-18 & DN-26\\
		\midrule
		\multirow{5}{*}{w/o SP}
		&PGD 			   &7.0\%    &13.9\%  & 12.3\%  & 19.1\% \\
		&DI  	       &6.9\%    &13.0\%  & 11.7\%  & 18.5\% \\
		&MI  		  &\textbf{{33.8\%}}    &\textbf{{50.9\%}}  & {45.3\%}   & {60.7\%} \\
		&TI-DIM         &10.0\%    &8.0\%  & 8.5\%  & 12.1\%\\
		\rowcolor {gray!20} \cellcolor{gray!0} &\textbf{ICE}    &21.9\%   & 23.6\%  & 23.3\%  & 35.1\% \\
		\midrule
		\multirow{6}{*}{Default}
		&PGD 			   &37.3\%    &50.9\%  & 43.2\%  & 61.5\%\\
		&DI 			   &39.3\%    &54.9\%  & 47.2\%  & 64.7\%\\
		&MI 			   &45.4\%    &62.3\%  & 56.3\%  & 72.1\%\\
		&TI-DIM 			   &51.0\%    &45.6\%  & 48.0\%  & 55.3\%\\
		\rowcolor {gray!20} \cellcolor{gray!0} &\textbf{ICE}    &{54.9\%}    &{67.3\%}  & {71.5\%}  & {83.5\%}\\
		\bottomrule
	\end{tabular}
	\label{tab:ablation}
\end{table}

\subsection{Ablation Study}
We verify the effect of each setting or component in our work via ablation experiments.
Note that the target dataset is Cifar-100 in all ablation experiments.

\subsubsection{The effect of SP}
\label{sec:adlation-SP}
When evaluating ICE and baselines on the GTA problem, we use the SP trick (the third step in Section \ref{sec:inference of ICE}) to improve their performances.
Here we validate the effectiveness of SP by removing it, which means for ICE, we directly use the adversarial example $x^{(T)}$ generated in the second step in Section \ref{sec:inference of ICE} to attack the target models.
Similarly, for the PGD-based baselines, we directly use $x_{adv}$ generated in the third step in Section \ref{sec:evaluate of baselines} to attack the target models.
Table~\ref{tab:ablation} reports the experimental results.
The source model and source dataset are RN-18 and Cifar-10, respectively.
It is clear that without SP, the performances of all methods are damaged.
The reason is that SP enlarges the average perturbation scale of each pixel, which possibly is an important factor for GTA performance.

\subsubsection{Smaller perturbation}
\label{sec:smaller_eps}
Here we test whether the proposed ICE and SA are sensitive to the perturbation scale by changing $\epsilon$ from the default 15 to 8.
The source datasets are Cifar-10, Tiered$_{T84}$, and Tiered$_{T56}$.
The source models are three RN-18 models respectively trained on the three source datasets.
Table~\ref{tab:more experiment on cifar100} shows the experimental results.
Both ICE and SA outperform baselines with significant margins.
In lots of cases, ICE and SA double the performances achieved by baselines, which further indicates the effectiveness of the proposed ICE and SA.

\begin{figure*}[t]
	\centering
	\includegraphics[width=0.95\textwidth]{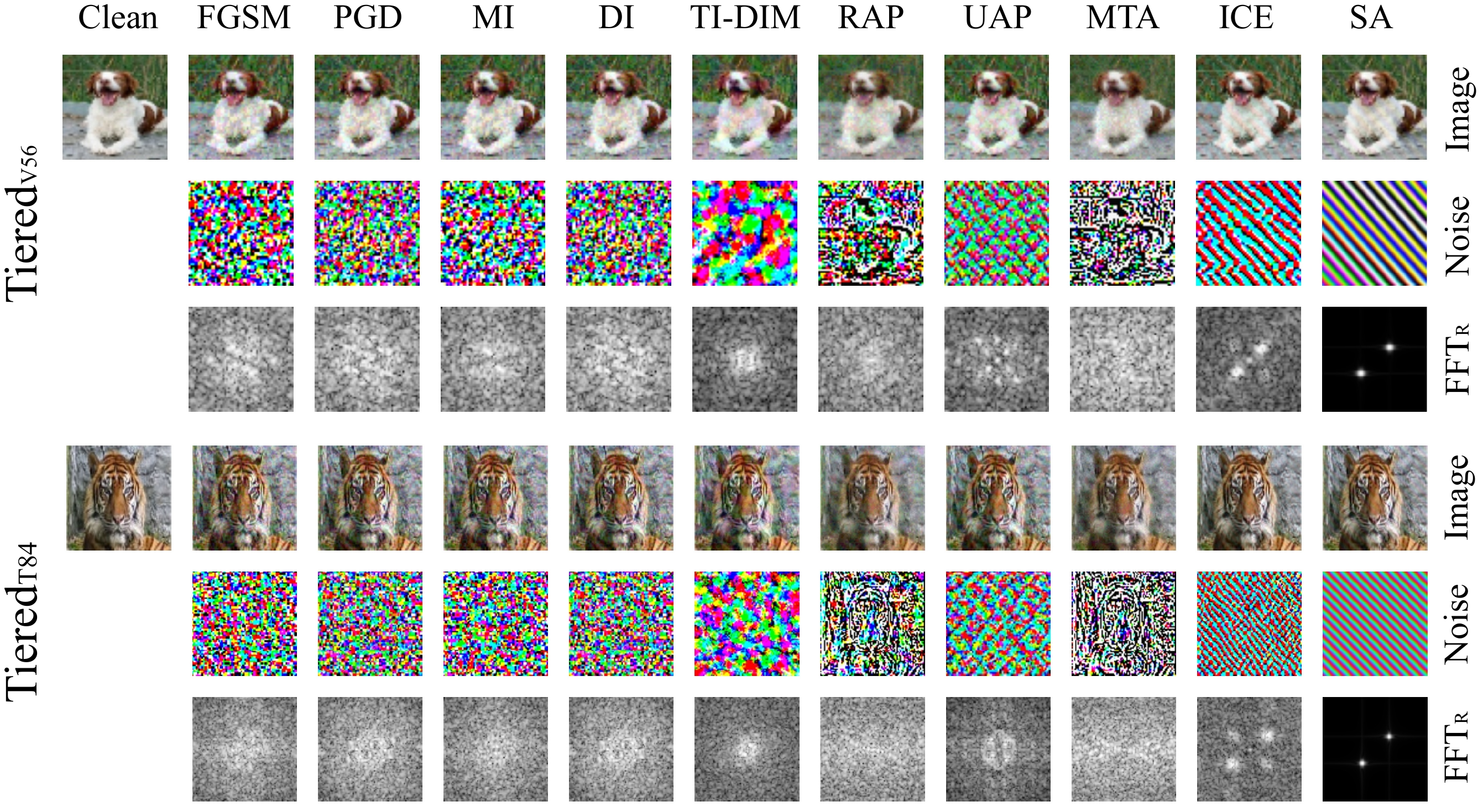}
	\caption{
		Some adversarial examples for clean images from Tiered$_{V56}$ and Tiered$_{T84}$.
		The adversarial examples are generated via FGSM, PGD, MI, DI, TI-DIM, RAP, UAP, MTA, ICE, and SA, with $\epsilon=15$.
		For visualization, we normalize the visualized noises into the range of 0 to 255.
	}
	\label{fig:visualization}
\end{figure*}

\subsection{Attacking object detection images}
\label{sec:attacking object detection}
We conduct an experiment to test whether the proposed ICE and SA can be directly used to perturb object detection images.
In this experiment, the target dataset is MS-COCO\cite{lin2014microsoft}, a popularly used object detection dataset, and the victim models are RetinaNet\cite{lin2017focal}, Faster-RCNN\cite{ren2015faster}, and Mask-RCNN\cite{he2017mask}.
All the victim models use ResNet-50\cite{he2016deep} as backbone.
The source datasets are Cifar-10, Tiered$_{T84}$, and Tiered$_{V56}$, and the source models are three RN-18 models respectively trained on the three source datasets.
$\epsilon$ is also set to 15.
Note that for each method (include baselines), we still perturb MS-COCO images with the same pipelines as we do in Sections~\ref{sec:attack cifar100}, \ref{sec:exp_attacking_across_datasets}, \ref{sec:attacking CUB}, and \ref{sec:attacking ImageNet} without any special modification (\emph{i.e.}, just treat MS-COCO images as classification images). 
For each MS-COCO image and each attack method, we create the adversarial image by the following steps: \textbf{1)} resize the clean image to 512$\times$512; \textbf{2)} use the attack method to perturb the resized MS-COCO image; \textbf{3)} resize the perturbed MS-COCO image back into the original shape.
The metric is average precision of predicted bounding boxes (AP$_{bbox}$).
Table~\ref{tab:attack_object detection} reports the experimental results.
Among all methods, SA performs the best to perturb MS-COCO images.
For instance, from RetinaNet's clean AP$_{bbox}$ 36.6\%, SA decreases AP$_{bbox}$ to 19.3\% while UAP can only decrease AP$_{bbox}$ to 22.5\%. 
We will visualize the perturbed MS-COCO images and the detection results in Appendix.

\begin{table}[t!]
	\centering
	\caption{GTA success rates on Cifar-100 with $\epsilon$=8.
	}
	{\footnotesize  
		\begin{tabular}{cccccc}
			\toprule
			Resource & Method & MN-V3 & VGG-16 & RN-18 & DN-26\\
			\midrule
			\multirow{8}{*}{\tabincell{c}{Cifar-10 \\ + Tiered$_{T84}$ \\ + Tiered$_{V56}$  \\  (RN-18)}}
			&FGSM 		      &18.3\%     &29.9\% & 22.3\%    & 39.7\% \\
			&MI 			   &17.5\%    &28.5\%  & 21.6\%  & 39.5\%\\
			&AEG  	     &21.1\%    &29.7\%  & 23.5\%  & 38.0\%\\
			&MTA  	     &17.2\%    &26.5\%  & 21.0\%  & 35.3\% \\
			&UAP  	     &20.9\%    &37.5\%  & 34.0\%  & 48.3\% \\
			&RAP  	     &15.2\%    &31.9\%  & 24.5\%  & 41.0\% \\
			\rowcolor {gray!20} \cellcolor{gray!0} &\textbf{ICE}    &{26.9\%}    &\textbf{49.9\%}  & \textbf{62.4\%}  & \textbf{72.2\%}\\
			\rowcolor {gray!20} \cellcolor{gray!0} &\textbf{SA}    &\textbf{36.0\%}    &{45.3\%}  & {53.9\%}  & {58.6\%}\\
			\bottomrule
		\end{tabular}
	}
	\label{tab:more experiment on cifar100}
\end{table}

\begin{table}[t]
	\centering
	\caption{GTA results on object detection images. }
	\setlength\tabcolsep{3.5pt}
	\scriptsize  
	\begin{tabular}{ccccc}
		\toprule
		Resource & Method & RetinaNet & Faster-RCNN & Mask-RCNN\\
		\midrule
		\multirow{10}{*}{\tabincell{c}{Cifar-10 \\ + Tiered$_{T84}$ \\ + Tiered$_{V56}$  \\  (RN-18)}}
		&Clean 		       &36.6\%    &37.4\%  & 38.0\% \\
		\midrule
		&FGSM 		       &34.0\%($\downarrow$2.6\%)    &34.7\%($\downarrow$2.7\%)  & 35.7\%($\downarrow$2.3\%) \\
		&MI 			   &34.2\%($\downarrow$2.4\%)    &34.9\%($\downarrow$2.5\%)  & 35.5\%($\downarrow$2.5\%) \\
		&TI-DIM 		   &22.9\%($\downarrow$13.7\%)    &23.0\%($\downarrow$14.4\%)  & 23.7\%($\downarrow$14.3\%) \\
		&MTA     		   &26.0\%($\downarrow$10.6\%)    &26.8\%($\downarrow$10.6\%)  & 27.5\%($\downarrow$10.5\%) \\
		&UAP     		   &22.5\%($\downarrow$14.1\%)    &23.2\%($\downarrow$14.2\%)  & 23.8\%($\downarrow$14.2\%) \\
		&RAP     		   &24.6\%($\downarrow$12.0\%)    &25.1\%($\downarrow$12.3\%)  & 25.9\%($\downarrow$12.1\%) \\
		\rowcolor {gray!20} \cellcolor{gray!0} &\textbf{ICE}    &{27.7\%($\downarrow$8.9\%)}    &{28.2\%($\downarrow$8.8\%)} & 28.8\%($\downarrow$9.2\%) \\
		\rowcolor {gray!20} \cellcolor{gray!0} &\textbf{SA}    &{19.3\%}\textbf{($\downarrow$17.3\%)}    &{19.7\%}\textbf{($\downarrow$17.7\%)} & {20.3\%}\textbf{($\downarrow$17.7\%)} \\
		\bottomrule
	\end{tabular}
	\label{tab:attack_object detection}
\end{table}

\section{Visualization}
\label{sec:visualization}
We visualize generated adversarial examples, the noise maps, and spectrum diagrams in Fig.\ref{fig:visualization}.
To save space, we only sample two clean images from Tiered$_{T84}$ and Tiered$_{V56}$, respectively, for visualization, and will show more visualizations in Appendix.
When attacking the sampled Tiered$_{T84}$ image, we use Cifar-10, Cifar-100, and Tiered$_{V56}$ as the source datasets, and use three RN-18 respectively trained on the source datasets as the source models.
When attacking the sampled Tiered$_{V56}$ image, we use Cifar-10, Cifar-100, and Tiered$_{T84}$ as the source datasets.
The raw FFT$_R$ denotes the spectrum of the noise map's R channel and Appendix will visualize FFT$_G$ and FFT$_B$.

By analyzing both Figs.\ref{fig:fig1} and \ref{fig:visualization}, we conclude that the perturbation noise generated by ICE differs greatly from those generated by the other methods.
Compared with other methods which use irregular noise patterns to perturb different images, ICE uses regular similar texture-like noises to perturb images from different datasets.
More visualizations in Appendix will also support this phenomenon.
The possible reason for this phenomenon is that ICE needs to disturb the input image without knowing any information in advance, which forces ICE to learn the perturbation pattern that correlates little with ground-truth information.
In our opinion, learning a generalizable perturbation pattern may be a straightforward way for ICE to solve GTA.

The spectrum diagrams demonstrate that the main components in each ICE noise are three sine waves for R, G, and B channels.
The proposed SA directly optimizes the three sine waves.
The visualization shows that the optimized sine waves use similar frequencies and directions with the three sine waves in ICE noise.

\section{Conclusion and future work}
In this paper, we propose the Generalized Transfer Attack (GTA) setting which is more challenging and more realistic than existing adversarial attack settings. We show that although existing attacks can be modified to solve this problem, they often perform poorly when the target dataset/model is different from sources. We then propose a novel Image Classification Eraser (ICE) method and show that ICE outperforms all the baselines in generalized transfer attack. 
More importantly, visualization and fast fourier transform analysis demonstrate that ICE uses similar texture-like noise pattern to perturb all images and the main components in each ICE noise are three sine waves.
Inspired by this finding, we propose another novel method Sine Attack which directly optimizes three sine waves to solve the GTA problem.
Experiments also demonstrate that SA outperforms comparably to ICE. 
Our work indicates that the three sine waves are enough and effective to perturb unknown images under the GTA setting.

{\small
	\bibliographystyle{ieee_fullname}
	\bibliography{main}

\begin{thebibliography}{10}\itemsep=-1pt

\bibitem{bose2020adversarial}
Avishek~Joey Bose, Gauthier Gidel, Hugo Berrard, Andre Cianflone, Pascal
  Vincent, Simon Lacoste-Julien, and William~L Hamilton.
\newblock Adversarial example games.
\newblock {\em Advances in neural information processing systems}, 2020.

\bibitem{brendel2018decision-based}
Wieland Brendel, Jonas Rauber, and Matthias Bethge.
\newblock Decision-based adversarial attacks: Reliable attacks against
  black-box machine learning models.
\newblock {\em international conference on learning representations}, 2018.

\bibitem{carlini2017towards}
Nicholas Carlini and David Wagner.
\newblock Towards evaluating the robustness of neural networks.
\newblock In {\em 2017 ieee symposium on security and privacy (sp)}, pages
  39--57. IEEE, 2017.

\bibitem{chen2017zoo}
Pin-Yu Chen, Huan Zhang, Yash Sharma, Jinfeng Yi, and Cho-Jui Hsieh.
\newblock Zoo: Zeroth order optimization based black-box attacks to deep neural
  networks without training substitute models.
\newblock In {\em Proceedings of the 10th ACM workshop on artificial
  intelligence and security}, pages 15--26, 2017.

\bibitem{cheng2018query}
Minhao Cheng, Thong Le, Pin-Yu Chen, Jinfeng Yi, Huan Zhang, and Cho-Jui Hsieh.
\newblock Query-efficient hard-label black-box attack: An optimization-based
  approach.
\newblock {\em arXiv preprint arXiv:1807.04457}, 2018.

\bibitem{cheng2020signopt}
Minhao Cheng, Simranjit Singh, Patrick~H. Chen, Pin-Yu Chen, Sijia Liu, and
  Cho-Jui Hsieh.
\newblock Sign-opt: A query-efficient hard-label adversarial attack.
\newblock In {\em international conference on learning representations}, 2020.

\bibitem{cheng2019improving}
Shuyu Cheng, Yinpeng Dong, Tianyu Pang, Hang Su, and Jun Zhu.
\newblock Improving black-box adversarial attacks with a transfer-based prior.
\newblock pages 10932--10942, 2019.

\bibitem{croce2020minimally}
Francesco Croce and Matthias Hein.
\newblock Minimally distorted adversarial examples with a fast adaptive
  boundary attack.
\newblock In {\em International Conference on Machine Learning}, pages
  2196--2205. PMLR, 2020.

\bibitem{Cubuk_2019_CVPR}
Ekin~D. Cubuk, Barret Zoph, Dandelion Mane, Vijay Vasudevan, and Quoc~V. Le.
\newblock Autoaugment: Learning augmentation strategies from data.
\newblock In {\em Proceedings of the IEEE/CVF Conference on Computer Vision and
  Pattern Recognition (CVPR)}, June 2019.

\bibitem{das2018shield}
Nilaksh Das, Madhuri Shanbhogue, Shang-Tse Chen, Fred Hohman, Siwei Li, Li
  Chen, Michael~E Kounavis, and Duen~Horng Chau.
\newblock Shield: Fast, practical defense and vaccination for deep learning
  using jpeg compression.
\newblock In {\em Proceedings of the 24th ACM SIGKDD International Conference
  on Knowledge Discovery \& Data Mining}, pages 196--204, 2018.

\bibitem{demontis2019why}
Ambra Demontis, Marco Melis, Maura Pintor, Matthew Jagielski, Battista Biggio,
  Alina Oprea, Cristina Nita-Rotaru, and Fabio Roli.
\newblock Why do adversarial attacks transfer? explaining transferability of
  evasion and poisoning attacks.
\newblock {\em USENIX Security Symposium}, pages 321--338, 2019.

\bibitem{deng2009imagenet}
Jia Deng, Wei Dong, Richard Socher, Li-Jia Li, Kai Li, and Li Fei-Fei.
\newblock Imagenet: A large-scale hierarchical image database.
\newblock In {\em IEEE conference on computer vision and pattern recognition},
  pages 248--255. Ieee, 2009.

\bibitem{dong2018boosting}
Yinpeng Dong, Fangzhou Liao, Tianyu Pang, Hang Su, Jun Zhu, Xiaolin Hu, and
  Jianguo Li.
\newblock Boosting adversarial attacks with momentum.
\newblock In {\em Proceedings of the IEEE conference on computer vision and
  pattern recognition}, 2018.

\bibitem{dong2019evading}
Yinpeng Dong, Tianyu Pang, Hang Su, and Jun Zhu.
\newblock Evading defenses to transferable adversarial examples by
  translation-invariant attacks.
\newblock In {\em Proceedings of the IEEE/CVF Conference on Computer Vision and
  Pattern Recognition}, pages 4312--4321, 2019.

\bibitem{du2020query-efficient}
Jiawei Du, Hu Zhang, Tianyi~Joey Zhou, Yi Yang, and Jiashi Feng.
\newblock Query-efficient meta attack to deep neural networks.
\newblock {\em International Conference on Learning Representations}, 2020.

\bibitem{finn2017model}
Chelsea Finn, Pieter Abbeel, and Sergey Levine.
\newblock Model-agnostic meta-learning for fast adaptation of deep networks.
\newblock In {\em International Conference on Machine Learning}, pages
  1126--1135. PMLR, 2017.

\bibitem{ganeshan2019fda}
Aditya Ganeshan, Vivek BS, and R~Venkatesh Babu.
\newblock Fda: Feature disruptive attack.
\newblock In {\em Proceedings of the IEEE/CVF International Conference on
  Computer Vision}, pages 8069--8079, 2019.

\bibitem{gao2020patch}
Lianli Gao, Qilong Zhang, Jingkuan Song, Xianglong Liu, and Heng~Tao Shen.
\newblock Patch-wise attack for fooling deep neural network.
\newblock In {\em European Conference on Computer Vision}, pages 307--322.
  Springer, 2020.

\bibitem{geirhos2019imagenet-trained}
Robert Geirhos, Patricia Rubisch, Claudio Michaelis, Matthias Bethge, A.~Felix
  Wichmann, and Wieland Brendel.
\newblock Imagenet-trained cnns are biased towards texture; increasing shape
  bias improves accuracy and robustness.
\newblock In {\em ICLR}, 2019.

\bibitem{goodfellow2014explaining}
Ian~J Goodfellow, Jonathon Shlens, and Christian Szegedy.
\newblock Explaining and harnessing adversarial examples.
\newblock {\em international conference on learning representations}, 2014.

\bibitem{guo2020backpropagating}
Yiwen Guo, Qizhang Li, and Hao Chen.
\newblock Backpropagating linearly improves transferability of adversarial
  examples.
\newblock In {\em Advances in neural information processing systems 33 (NIPS
  2020)}, 2020.

\bibitem{he2017mask}
Kaiming He, Georgia Gkioxari, Piotr Doll{\'a}r, and Ross Girshick.
\newblock Mask r-cnn.
\newblock In {\em Proceedings of the IEEE international conference on computer
  vision}, pages 2961--2969, 2017.

\bibitem{he2016deep}
Kaiming He, Xiangyu Zhang, Shaoqing Ren, and Jian Sun.
\newblock Deep residual learning for image recognition.
\newblock In {\em Proceedings of the IEEE conference on computer vision and
  pattern recognition}, pages 770--778, 2016.

\bibitem{howard2019searching}
Andrew Howard, Mark Sandler, Grace Chu, Liang-Chieh Chen, Bo Chen, Mingxing
  Tan, Weijun Wang, Yukun Zhu, Ruoming Pang, Vijay Vasudevan, et~al.
\newblock Searching for mobilenetv3.
\newblock In {\em Proceedings of the IEEE/CVF International Conference on
  Computer Vision}, pages 1314--1324, 2019.

\bibitem{huang2017densely}
Gao Huang, Zhuang Liu, Laurens Van Der~Maaten, and Kilian~Q Weinberger.
\newblock Densely connected convolutional networks.
\newblock In {\em Proceedings of the IEEE conference on computer vision and
  pattern recognition}, pages 4700--4708, 2017.

\bibitem{huang2019enhancing}
Qian Huang, Isay Katsman, Horace He, Zeqi Gu, Serge Belongie, and Ser-Nam Lim.
\newblock Enhancing adversarial example transferability with an intermediate
  level attack.
\newblock In {\em Proceedings of the IEEE/CVF International Conference on
  Computer Vision}, pages 4733--4742, 2019.

\bibitem{huang2020black}
Zhichao Huang and Tong Zhang.
\newblock Black-box adversarial attack with transferable model-based embedding.
\newblock {\em International Conference on Learning Representations}, 2020.

\bibitem{ilyas2018black}
Andrew Ilyas, Logan Engstrom, Anish Athalye, and Jessy Lin.
\newblock Black-box adversarial attacks with limited queries and information.
\newblock In {\em International Conference on Machine Learning}, pages
  2137--2146. PMLR, 2018.

\bibitem{ilyas2019adversarial}
Andrew Ilyas, Shibani Santurkar, Dimitris Tsipras, Logan Engstrom, Brandon
  Tran, and Aleksander Madry.
\newblock Adversarial examples are not bugs, they are features.
\newblock {\em arXiv preprint arXiv:1905.02175}, 2019.

\bibitem{inkawhich2021can}
Nathan Inkawhich, Kevin~J Liang, Jingyang Zhang, Huanrui Yang, Hai Li, and
  Yiran Chen.
\newblock Can targeted adversarial examples transfer when the source and target
  models have no label space overlap?
\newblock {\em arXiv preprint arXiv:2103.09916}, 2021.

\bibitem{kaidi2019structured}
Xu Kaidi, Liu Sijia, Zhao Pu, Chen Pin-Yu, Zhang Huan, Fan Quanfu, Erdogmus
  Deniz, Wang Yanzhi, and Lin Xue.
\newblock Structured adversarial attack: Towards general implementation and
  better interpretability.
\newblock {\em International Conference on Learning Representations}, 2019.

\bibitem{krizhevsky2009learning}
Alex Krizhevsky, Geoffrey Hinton, et~al.
\newblock Learning multiple layers of features from tiny images.
\newblock 2009.

\bibitem{kurakin2016adversarial}
Alexey Kurakin, Ian Goodfellow, Samy Bengio, et~al.
\newblock Adversarial examples in the physical world, 2016.

\bibitem{li2020qeba}
Huichen Li, Xiaojun Xu, Xiaolu Zhang, Shuang Yang, and Bo Li.
\newblock Qeba: Query-efficient boundary-based blackbox attack.
\newblock In {\em Proceedings of the IEEE/CVF Conference on Computer Vision and
  Pattern Recognition}, pages 1221--1230, 2020.

\bibitem{li2020practical}
Qizhang Li, Yiwen Guo, and Hao Chen.
\newblock Practical no-box adversarial attacks against dnns.
\newblock {\em Advances In Neural Information Processing Systems 2020}, 2020.

\bibitem{li2020learning}
Yingwei Li, Song Bai, Yuyin Zhou, Cihang Xie, Zhishuai Zhang, and Alan Yuille.
\newblock Learning transferable adversarial examples via ghost networks.
\newblock {\em AAAI}, pages 11458--11465, 2020.

\bibitem{lin2020nesterov}
Jiadong Lin, Chuanbiao Song, Kun He, Liwei Wang, and John~E Hopcroft.
\newblock Nesterov accelerated gradient and scale invariance for adversarial
  attacks.
\newblock {\em International Conference on Learning Representations}, 2020.

\bibitem{lin2017focal}
Tsung-Yi Lin, Priya Goyal, Ross Girshick, Kaiming He, and Piotr Doll{\'a}r.
\newblock Focal loss for dense object detection.
\newblock In {\em Proceedings of the IEEE international conference on computer
  vision}, pages 2980--2988, 2017.

\bibitem{lin2014microsoft}
Tsung-Yi Lin, Michael Maire, Serge Belongie, James Hays, Pietro Perona, Deva
  Ramanan, Piotr Doll{\'a}r, and C~Lawrence Zitnick.
\newblock Microsoft coco: Common objects in context.
\newblock In {\em European conference on computer vision}, pages 740--755.
  Springer, 2014.

\bibitem{liu2019darts}
Hanxiao Liu, Karen Simonyan, and Yiming Yang.
\newblock Darts: Differentiable architecture search.
\newblock In {\em international conference on learning representations}, 2019.

\bibitem{liu2016delving}
Yanpei Liu, Xinyun Chen, Chang Liu, and Dawn Song.
\newblock Delving into transferable adversarial examples and black-box attacks.
\newblock {\em international conference on learning representations}, 2017.

\bibitem{liu2019metapruning}
Zechun Liu, Haoyuan Mu, Xiangyu Zhang, Zichao Guo, Xin Yang, Kwang-Ting Cheng,
  and Jian Sun.
\newblock Metapruning: Meta learning for automatic neural network channel
  pruning.
\newblock In {\em Proceedings of the IEEE/CVF International Conference on
  Computer Vision}, pages 3296--3305, 2019.

\bibitem{Lu_2020_CVPR}
Yantao Lu, Yunhan Jia, Jianyu Wang, Bai Li, Weiheng Chai, Lawrence Carin, and
  Senem Velipasalar.
\newblock Enhancing cross-task black-box transferability of adversarial
  examples with dispersion reduction.
\newblock In {\em Proceedings of the IEEE/CVF Conference on Computer Vision and
  Pattern Recognition (CVPR)}, June 2020.

\bibitem{maksym2020square}
Andriushchenko Maksym, Croce Francesco, Flammarion Nicolas, and Hein Matthias.
\newblock Square attack: a query-efficient black-box adversarial attack via
  random search.
\newblock {\em european conference on computer vision}, pages 484--501, 2020.

\bibitem{SNAIL}
Nikhil Mishra, Mostafa Rohaninejad, Xi Chen, and Pieter Abbeel.
\newblock A simple neural attentive meta-learner.
\newblock {\em international conference on learning representations}, 2018.

\bibitem{moosavi2017universal}
Seyed-Mohsen Moosavi-Dezfooli, Alhussein Fawzi, Omar Fawzi, and Pascal
  Frossard.
\newblock Universal adversarial perturbations.
\newblock In {\em Proceedings of the IEEE conference on computer vision and
  pattern recognition}, 2017.

\bibitem{moosavi2016deepfool}
Seyed-Mohsen Moosavi-Dezfooli, Alhussein Fawzi, and Pascal Frossard.
\newblock Deepfool: a simple and accurate method to fool deep neural networks.
\newblock In {\em Proceedings of the IEEE conference on computer vision and
  pattern recognition}, pages 2574--2582, 2016.

\bibitem{naseer2019cross-domain}
Muhammad~Muzammal Naseer, Salman~H Khan, Muhammad~Haris Khan, Fahad
  Shahbaz~Khan, and Fatih Porikli.
\newblock Cross-domain transferability of adversarial perturbations.
\newblock {\em Advances in Neural Information Processing Systems},
  32:12905--12915, 2019.

\bibitem{nathan2020perturbing}
Inkawhich Nathan, Kevin~Liang J, Wang Binghui, Inkawhich Matthew, Carin
  Lawrence, and Chen Yiran.
\newblock Perturbing across the feature hierarchy to improve standard and
  strict blackbox attack transferability.
\newblock In {\em NIPS}, 2020.

\bibitem{papernot2016transferability}
Nicolas Papernot, Patrick McDaniel, and Ian Goodfellow.
\newblock Transferability in machine learning: from phenomena to black-box
  attacks using adversarial samples.
\newblock {\em arXiv preprint arXiv:1605.07277}, 2016.

\bibitem{papernot2017practical}
Nicolas Papernot, Patrick McDaniel, Ian Goodfellow, Somesh Jha, Z~Berkay Celik,
  and Ananthram Swami.
\newblock Practical black-box attacks against machine learning.
\newblock In {\em Proceedings of the 2017 ACM on Asia conference on computer
  and communications security}, pages 506--519, 2017.

\bibitem{prakash2021multi}
Aditya Prakash, Kashyap Chitta, and Andreas Geiger.
\newblock Multi-modal fusion transformer for end-to-end autonomous driving.
\newblock In {\em Proceedings of the IEEE/CVF Conference on Computer Vision and
  Pattern Recognition}, pages 7077--7087, 2021.

\bibitem{qin2021training}
Yunxiao Qin, Yuanhao Xiong, Jinfeng Yi, and Cho-Jui Hsieh.
\newblock Training meta-surrogate model for transferable adversarial attack.
\newblock {\em arXiv preprint arXiv:2109.01983}, 2021.

\bibitem{meta-teacher}
Yunxiao Qin, Zitong Yu, Longbin Yan, Zezheng Wang, Chenxu Zhao, and Zhen Lei.
\newblock Meta-teacher for face anti-spoofing.
\newblock {\em IEEE Transactions on Pattern Analysis and Machine Intelligence},
  pages 1--1, 2021.

\bibitem{qin2020layer}
Yunxiao Qin, Weiguo Zhang, Zezheng Wang, Chenxu Zhao, and Jingping Shi.
\newblock Layer-wise adaptive updating for few-shot image classification.
\newblock {\em IEEE Signal Processing Letters}, 27:2044--2048, 2020.

\bibitem{ren2018meta}
Mengye Ren, Eleni Triantafillou, Sachin Ravi, Jake Snell, Kevin Swersky,
  Joshua~B Tenenbaum, Hugo Larochelle, and Richard~S Zemel.
\newblock Meta-learning for semi-supervised few-shot classification.
\newblock {\em International Conference on Learning Representations}, 2018.

\bibitem{ren2018learning}
Mengye Ren, Wenyuan Zeng, Bin Yang, and Raquel Urtasun.
\newblock Learning to reweight examples for robust deep learning.
\newblock In {\em International Conference on Machine Learning}, pages
  4334--4343. PMLR, 2018.

\bibitem{ren2015faster}
Shaoqing Ren, Kaiming He, Ross Girshick, and Jian Sun.
\newblock Faster r-cnn: towards real-time object detection with region proposal
  networks.
\newblock {\em IEEE transactions on pattern analysis and machine intelligence},
  39(6):1137--1149, 2016.

\bibitem{simonyan2015very}
Karen Simonyan and Andrew Zisserman.
\newblock Very deep convolutional networks for large-scale image recognition.
\newblock {\em International Conference on Learning Representations}, 2015.

\bibitem{szegedy2014intriguing}
Christian Szegedy, Wojciech Zaremba, Ilya Sutskever, Joan Bruna, Dumitru Erhan,
  J.~Ian Goodfellow, and Rob Fergus.
\newblock Intriguing properties of neural networks.
\newblock {\em international conference on learning representations}, 2014.

\bibitem{tramer2017ensemble}
Florian Tram{\`e}r, Alexey Kurakin, Nicolas Papernot, Ian Goodfellow, Dan
  Boneh, and Patrick McDaniel.
\newblock Ensemble adversarial training: Attacks and defenses.
\newblock {\em arXiv preprint arXiv:1705.07204}, 2017.

\bibitem{WahCUB_200_2011}
C. Wah, S. Branson, P. Welinder, P. Perona, and S. Belongie.
\newblock {The Caltech-UCSD Birds-200-2011 Dataset}.
\newblock Technical Report CNS-TR-2011-001, California Institute of Technology,
  2011.

\bibitem{wang2020spanning}
Lu Wang, Huan Zhang, Jinfeng Yi, Cho-Jui Hsieh, and Yuan Jiang.
\newblock Spanning attack: reinforce black-box attacks with unlabeled data.
\newblock {\em Machine Learning}, 109(12):2349--2368, 2020.

\bibitem{wang2021enhancing}
Xiaosen Wang and Kun He.
\newblock Enhancing the transferability of adversarial attacks through variance
  tuning.
\newblock In {\em Proceedings of the IEEE/CVF Conference on Computer Vision and
  Pattern Recognition}, pages 1924--1933, 2021.

\bibitem{wang2021unified}
Xin Wang, Jie Ren, Shuyun Lin, Xiangming Zhu, Yisen Wang, and Quanshi Zhang.
\newblock A unified approach to interpreting and boosting adversarial
  transferability.
\newblock {\em International Conference on Learning Representations}, 2021.

\bibitem{wu2020skip}
Dongxian Wu, Yisen Wang, Shu-Tao Xia, James Bailey, and Xingjun Ma.
\newblock Skip connections matter: On the transferability of adversarial
  examples generated with resnets.
\newblock {\em international conference on learning representations}, 2020.

\bibitem{wu2020stronger}
Kaiwen Wu, Allen Wang, and Yaoliang Yu.
\newblock Stronger and faster wasserstein adversarial attacks.
\newblock {\em International Conference on Machine Learning}, pages
  10377--10387, 2020.

\bibitem{wu2018understanding}
Lei Wu, Zhanxing Zhu, Cheng Tai, et~al.
\newblock Understanding and enhancing the transferability of adversarial
  examples.
\newblock {\em arXiv preprint arXiv:1802.09707}, 2018.

\bibitem{wu2020boosting}
Weibin Wu, Yuxin Su, Xixian Chen, Shenglin Zhao, Irwin King, R.~Michael Lyu,
  and Yu-Wing Tai.
\newblock Boosting the transferability of adversarial samples via attention.
\newblock In {\em Proceedings of the IEEE/CVF Conference on Computer Vision and
  Pattern Recognition}, pages 1158--1167, 2020.

\bibitem{xiao2021graph}
Zang Xiao, Xie Yi, Chen Jie, and Yuan Bo.
\newblock Graph universal adversarial attacks - a few bad actors ruin graph
  learning models.
\newblock {\em International Joint Conference on Artificial Intelligence},
  pages 3328--3334, 2021.

\bibitem{xie2019improving}
Cihang Xie, Zhishuai Zhang, Yuyin Zhou, Song Bai, Jianyu Wang, Zhou Ren, and
  Alan~L Yuille.
\newblock Improving transferability of adversarial examples with input
  diversity.
\newblock In {\em Proceedings of the IEEE/CVF Conference on Computer Vision and
  Pattern Recognition}, pages 2730--2739, 2019.

\bibitem{yin2019fourier}
Dong Yin, Raphael Gontijo~Lopes, Jon Shlens, Ekin~Dogus Cubuk, and Justin
  Gilmer.
\newblock A fourier perspective on model robustness in computer vision.
\newblock In {\em Advances in Neural Information Processing Systems},
  volume~32, 2019.

\bibitem{yuan2021meta}
Zheng Yuan, Jie Zhang, Yunpei Jia, Chuanqi Tan, Tao Xue, and Shiguang Shan.
\newblock Meta gradient adversarial attack.
\newblock In {\em Proceedings of the IEEE/CVF International Conference on
  Computer Vision}, 2021.

\bibitem{zhang2021data}
Chaoning Zhang, Philipp Benz, Adil Karjauv, and In~So Kweon.
\newblock Data-free universal adversarial perturbation and black-box attack.
\newblock In {\em Proceedings of the IEEE/CVF International Conference on
  Computer Vision}, pages 7868--7877, 2021.

\bibitem{zhang2021universal}
Chaoning Zhang, Philipp Benz, Adil Karjauv, and In~So Kweon.
\newblock Universal adversarial perturbations through the lens of deep
  steganography: Towards a fourier perspective.
\newblock In {\em Proceedings of the AAAI Conference on Artificial
  Intelligence}, volume~35, pages 3296--3304, 2021.

\bibitem{zhang2019bridging}
Wen Zhang, Yang Feng, Fandong Meng, Di You, and Qun Liu.
\newblock Bridging the gap between training and inference for neural machine
  translation.
\newblock In {\em 57th annual meeting of the association for computational
  linguistics (ACL 2019)}, pages 4334--4343, 2019.

\bibitem{zhou2018transferable}
Wen Zhou, Xin Hou, Yongjun Chen, Mengyun Tang, Xiangqi Huang, Xiang Gan, and
  Yong Yang.
\newblock Transferable adversarial perturbations.
\newblock In {\em Proceedings of the European Conference on Computer Vision
  (ECCV)}, pages 452--467, 2018.

\end{thebibliography}
}

\clearpage
\appendix
\section{Network architecture of ICE}
\label{sec:appendix_network}
\begin{figure*}[t]
	\centering
	\includegraphics[width=0.99\textwidth]{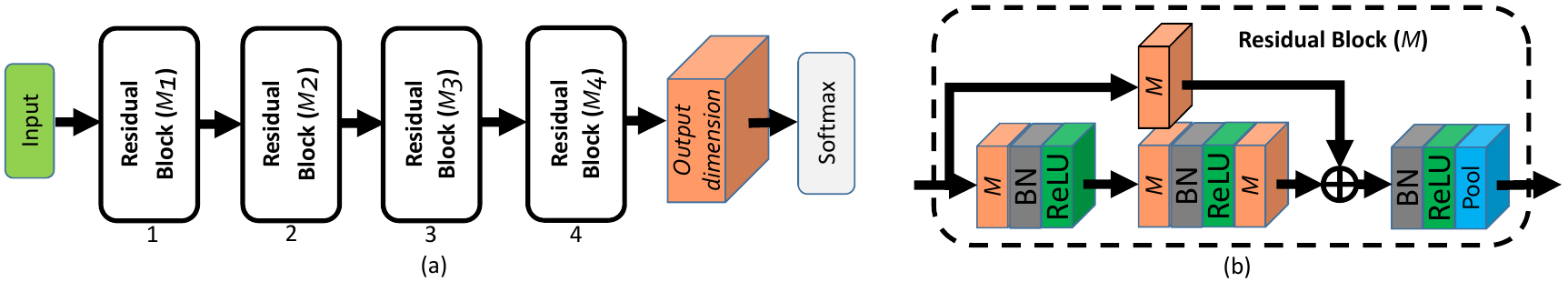}
	\caption{
		(a) The network architecture of the proposed ICE. 
		It is composed of four cascaded residual blocks, one convolutional layer, and one softmax layer.
		(b) The inner structure of the residual block.
		Orange cube denotes convolutional layer and the number on it denotes the number of filters of the convolutional layer.
		`Pool' in the last residual block is global average pooling and `Pool' in all the other residual blocks are max-pooling with both stride and pooling size set to 2.
	}
	\label{fig:network}
\end{figure*}

As mentioned in Section 4 of the main body, the proposed ICE should be able to handle images with different resolutions because we cannot know the image shape in advance.
Therefore, we build a fully convolutional neural network shown in Figure~\ref{fig:network} as the backbone of ICE.
The parameters $M_1$, $M_2$, $M_3$, and $M_4$ are set to 32, 64, 128, and 256, respectively.

\section{More details of Tiered$_{T84}$ and Tiered$_{V56}$}
\label{sec:appendix_tiered}
For either Tiered$_{T84}$ or Tiered$_{V56}$, we use the image ID to rank all images of each category and use the first 1200 images of each category to compose the training set and use the last 100 images to compose the testing set. 
For example, the IDs of the two images `n01530575\_5.JPEG' and `n01530575\_23.JPEG' from the `n01530575' category are 5 and 23.

\section{Training details of source and target models}
\label{sec:appendix_source_target_models}
On each of the datasets Cifar-10, Cifar-100, Tiered$_{T84}$, and Tiered$_{V56}$, we train the four models RN-18, VGG-16, MB-V3, and DN-26.
The network architectures of all the four models are defined in the public GitHub repository\footnote{https://github.com/yxlijun/cifar-tensorflow}.
We use consistent hyper-parameters to train all the models for 80,000 iterations without data augmentation.
The learning rate, L2 weight decay, and batch size are set to 0.01, 1e-5, and 128, respectively.
Table~\ref{tab:source and target models} shows the four models' accuracies on the four datasets.

\begin{table}[h]
	\centering
	\caption{Accuracies of source and target model on the four datasets.
	}
	{\scriptsize   
		\begin{tabular}{ccccc}
			\toprule
			Dataset & MB-V3 & VGG-16 & RN-18 & DN-26\\
			\midrule
			Cifar-10        &80.0\%    &92.9\%  & 91.8\%  & 91.2\%\\
			Cifar-100       &43.9\%   &68.5\%  & 68.0\%   & 64.6\%\\
			Tiered$_{T84}$ &35.3\%    &46.9\%  & 47.0\%   & 44.6\%\\
			Tiered$_{V56}$ &32.1\%  	  &46.3\%  & 45.9\%   & 46.0\%\\
			\bottomrule
		\end{tabular}
	}
	\label{tab:source and target models}
\end{table}

\section{Additional experiments}

\begin{figure}[t]
	\centering
	\includegraphics[width=0.95\linewidth]{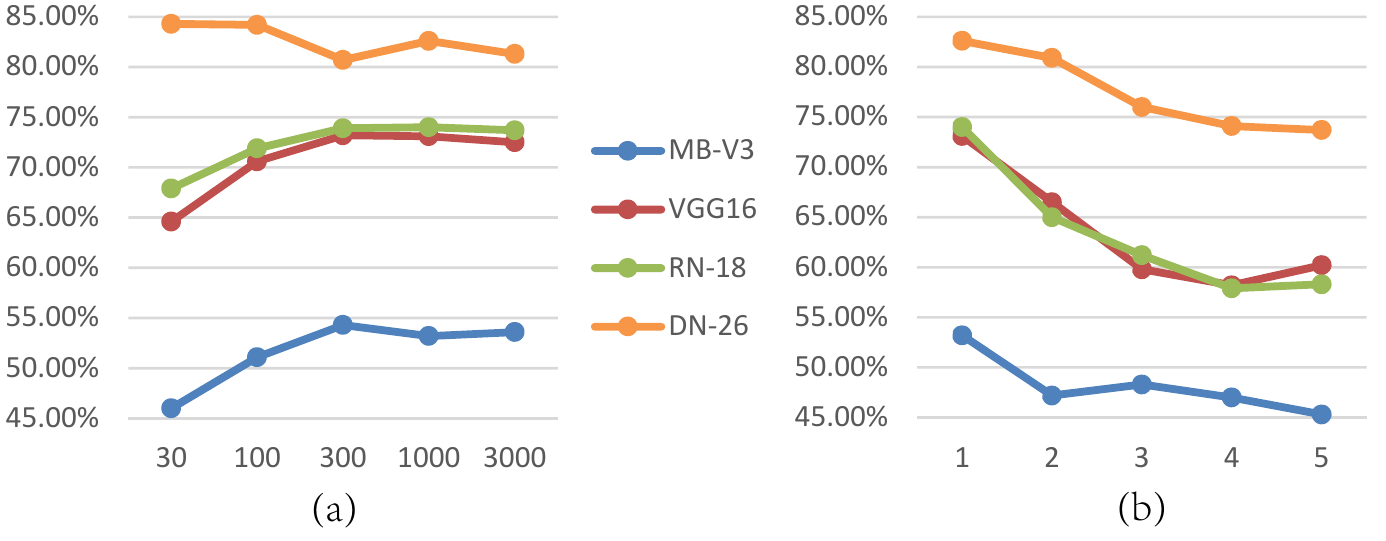}
	\caption{
		All experiments in this figure use Cifar-10 and ResNet-18 as the source dataset and the source model.
		(a): ICE's GTA results on four target models trained on Cifar-100 with different output dimensions.
		(b): ICE's GTA results on four target models trained on Cifar-100 with different numbers of T in the training phase.
	}
	\label{fig:ablation}
\end{figure}

\subsection{Output dimension of ICE}
We set the output dimension of ICE as 1,000 by default.
Here we show how the output dimension affects the results in Figure~\ref{fig:ablation}(a).
We can see that the average attack success rates on the four target models will rise when the output dimension is increased from 30 to 300, and will become stable when the output dimension $>$ 300.

\subsection{Customized FGSM or Customized PGD?}
\label{sec:ablation customized FGSM}
Eq.2 in the main body shows that we use Customized FGSM to perturb the input image in the inner-loop.
We conduct an experiment to show why we use Customized FGSM instead of Customized PGD~\cite{qin2021training}.
Customized PGD is a multi-step Customized FGSM. 
In this experiment, we increase the number of gradient ascent steps from 1 to 5.
Figure~\ref{fig:ablation}(b) shows the experimental results.
It is clear that the performance of ICE will be damaged by the increase of the number of gradient ascent steps.
The possible reason for this phenomenon is that multi-step gradient ascent in the inner-loop makes the model $\mathcal{U}_\theta$ hard to be optimized in the outer-loop.

\subsection{Attacking robust models}
\label{sec:attacking robust models}
Here we conduct generalized transfer attack on robust models.
In this experiment, we use RN-18 and Cifar-10 as the source model and the source dataset, and disturb the images from Cifar-100.
The target models are adversarially trained RN-18 and RN-34 on Cifar-100, which can be denoted as RN-18$_{adv}$ and RN-34$_{adv}$, respectively.
To obtain RN-18$_{adv}$, we firstly use the normally trained RN-18 to generate adversarial examples for all training examples with FGSM ($\epsilon$=15), and then retrain RN-18 on all the clean training images and the adversarial images.
We obtain RN-34$_{adv}$ in a similar way.
The two models finally achieve approximately 46.9\% and 47.3\% testing accuracies on Cifar-100.
The generalized transfer attack results on RN-18$_{adv}$ and RN-34$_{adv}$ are reported in Table \ref{tab:attack robust models}.  
Obviously, though all the methods perform not well to attack adversarially trained models, ICE and SA still show their advantages in this experiment.

\subsection{Re-implement PGD-based baselines with KL divergence}
\label{sec:KL divergence}
In the main body, the PGD-based baselines disturb input images by maximizing the entropy loss of source models.
Here we re-implement PGD-based baselines with KL divergence, which means the baselines disturb the input images by maximizing the KL divergence between the predicted distribution $\hat{y}$ and $\hat{y_0}$, where $\hat{y_0}$ is the predicted distribution for the original clean image $x$.
Then the perturbation generated in each gradient ascent step can be reformulated as $\delta= \text{sign}\big (\nabla_{x^{(i-1)}}\text{KL}(\hat{y}, \hat{y_0}) \big)$.
In this experiment, we use RN-18 and Cifar-10 as the source model and source dataset, and use Cifar-100 as the target dataset.
The generalized transfer attack results on MB-V3, VGG-16, RN-18, and DN-26 are reported in Table~\ref{tab:KL on cifar100}. 
The comparison between the results here and those in Table 2 demonstrates that the KL divergence cannot improve PGD-based baselines.

\begin{table}[]
	\centering
	\footnotesize 
	\caption{GTA success rates on robust Cifar-100 models. } 
	\begin{tabular}{cccc}
		\toprule
		Resource & Method & RN-18$_{adv}$ & RN-34$_{adv}$ \\
		\midrule
		\multirow{8}{*}{\tabincell{c}{Cifar-10  \\  (RN-18)}}
		&FGSM 		      &11.1\%     &12.9\% \\
		&PGD 			   &9.7\%    &10.1\%  \\
		&DI 			   &11.0\%    &11.6\% \\
		&MI 			   &10.8\%    &11.8\% \\
		&TI-DIM 		  &16.7\%    &15.9\% \\
		&IR         &13.8\%    &13.1\% \\
		&UAP         &11.2\%    &11.8\% \\
		&RAP         &14.9\%    &16.0\% \\
		&\textbf{ICE}    &\textbf{17.3\%}    &{17.8\%} \\
		&\textbf{SA}    &{16.0\%}    &\textbf{18.8\%} \\
		\bottomrule
	\end{tabular}
	\label{tab:attack robust models}
\end{table}

\begin{table}[h!]
	\centering
	\caption{GTA success rates on Cifar-100 with KL divergence. Target dataset is Cifar-100.
	}
	{\scriptsize  
		\begin{tabular}{cccccc}
			\toprule
			Resource & Method & MB-V3 & VGG-16 & RN-18 & DN-26\\
			\midrule
			\multirow{5}{*}{\tabincell{c}{Cifar-10 \\  (RN-18)}}
			&FGSM 		      &49.3\%     &63.3\% & 57.9\%  & 72.5\% \\
			&PGD 			   &39.6\%    &52.5\%  & 45.0\%  & 62.7\%\\
			&DI 			   &41.2\%    &54.6\%  & 46.5\%  & 64.2\%\\
			&MI 			   &47.3\%    &62.6\%  & 56.9\%  & 72.4\%\\
			&TI-DIM 		  &50.6\%    &43.1\%  & 45.6\%  & 52.4\%\\
			\bottomrule
		\end{tabular}
	}
	\label{tab:KL on cifar100}
\end{table}

\begin{table}[]
	\centering
	\caption{GTA success rates of the PGD-based baselines re-implemented with new pipeline (optimizing a single adversarial image). Target dataset is Cifar-100.
	}
	{\scriptsize  
		\begin{tabular}{cccccc}
			\toprule
			Resource & Method & MB-V3 & VGG-16 & RN-18 & DN-26\\
			\midrule
			\multirow{5}{*}{\tabincell{c}{Cifar-10 \\ + Tiered$_{T84}$ \\ + Tiered$_{V56}$  \\  (RN-18)}}
			&FGSM 		      &47.5\%     &63.2\% & 56.3\%    & 72.6\% \\
			&PGD 			   &38.8\%    &53.6\%  & 43.1\%  & 64.6\%\\
			&DI 			   &38.6\%    &53.9\%  & 43.3\%  & 64.7\%\\
			&MI 			   &46.2\%    &62.8\%  & 53.8\%   & 71.9\%\\
			&TI-DIM 		  &47.9\%    &43.8\%  & 44.9\%   & 54.0\%\\
			\bottomrule
		\end{tabular}
	}
	\label{tab:Optimize single image}
\end{table}

\subsection{Optimizing a single adversarial image}
\label{sec:optimize single}
In our previous experiments, we implement PGD-based baselines with the pipeline introduced in Section 6.1, where we firstly optimize the adversarial example for each source model respectively and then obtain the final adversarial example by fusing all the adversarial examples.
Here we re-implement PGD-based baselines with a new pipeline, where we directly optimize a single adversarial image for all source models.
The new pipeline uses $T$ gradient ascent iterations to disturb the input image, and the $i$-th iteration contains the following three steps.

\textbf{1)} Suppose $x^{(i-1)}$ is the perturbed image generated in the $i-1$ iteration and $x^{(0)}=x$.
We resize the image $x^{(i-1)}$ to the input shapes of all source models and then feed the resized images to the source models, respectively. 
Then the inference of $x^{(i-1)}$ on one source model $\mathbf{M}_{D_k}^j$ can be formulated as $y_{D_k}^j = \mathbf{M}_{D_k}^j(x_{D_k}^j)$, where $x_{D_k}^j = \text{resize}(x^{(i-1)}, \text{resolution}(\mathbf{M}_{D_k}^j))$ is the resized image for the source model $\mathbf{M}_{D_k}^j$. 

\textbf{2)} We calculate the prediction losses of all the source models with the formulation $L = \frac{1}{m} \cdot \sum_{k=1}^{m} \cdot \big ( \frac{1}{N_k} \cdot  \sum_{j=1}^{N_k}\mathcal{L}(y_{D_k}^{j}) \big)$, where $\mathcal{L}$ is entropy.  

\textbf{3)} We disturb the input image $x^{(i-1)}$ with the formulation $x^{(i)} =  \text{clip}(x^{(i-1)} + \frac{\epsilon}{T} \cdot \delta )$, where $	\delta = \text{sign}\big (\nabla_{x^{(i-1)}}{L} \big )$.

After $T$ iterations, we obtain the disturbed image $x^{(T)}$, and generate the final adversarial image $\hat{x}$ by $\hat{x} = \text{clip}(x+\epsilon \cdot \text{sign}(x^{(T)} - x))$. 
Then we feed both $\hat{x}$ and the clean image $x$ into the unknown target model $\mathbb{M}$ and get the predictions.
The GTA process is successful if $\mathbb{M}(\hat{x}) \neq \mathbb{M}(x)$.

We conduct an experiment to evaluate how the re-implemented PGD-based baselines perform on GTA.
In this experiment, Cifar-10, Tiered$_{T84}$, and Tiered$_{V56}$ are used as source datasets, and three RN-18 respectively trained on the three datasets are used as the source models.
The experimental results on the target models MB-V3, VGG-16, RN-18, and DN-26 are reported in Table~\ref{tab:Optimize single image}.
Compared with the results in the last row of Table 2, optimizing a single image improves FGSM and MI, but damages DI and TI-DIM.

The possible reason for this result is that \textbf{there still exists implicit resizing operations when we optimize a single adversarial image (the gradient propagates through resizing layers) at each gradient ascent step.}

Suppose there are three source models respectively trained on Tiered$_{T84}$, Tiered$_{V56}$, and Cifar-10. 
With the three source datasets, each gradient ascent step needs 3 explicit and 3 implicit resizing operations in the forward and backward paths, respectively.
Optimizing a single adversarial image needs to fuse the gradients from all the three source models at each gradient ascent step. 
This may result in the single adversarial example obtained in the step being not optima and accurate to attack each source model due to the implicit resizing operations, and further influences the following gradient ascent steps.
As introduced in Section 6.1, \textbf{the default implementation of baselines explicitly resizes the adversarial images only before the first and only after the last gradient ascent steps. }
For example, when using the Cifar-10 model to generate the perturbation, we explicitly resize the image $x$ to the resolution of Cifar-10 model only before the first gradient ascent step, and explicitly resize the adversarial example on the Cifar-10 model back into the original resolution of $x$ only after the last gradient ascent step. 
Therefore, between the first and the last step, there are no resizing operations to influence the adversarial effectiveness of the adversarial examples generated on the Cifar-10, Tiered$_{T84}$, and Tiered$_{V56}$ models, respectively.

The experiment shows that FGSM can be improved by optimizing a single adversarial image, which may prove the above reason. FGSM contains a single gradient ascent step, which means the first and the last gradient ascent steps is the same. \textbf{When optimizing a single adversarial image, FGSM has 3 implicit and 3 explicit resizing operations, while when using the default implementation introduced in Section 6.1, FGSM contains 3$\times$2=6 explicit resizing operations.} In our opinion, gradient fusing through implicit resizing operations is more accurate than through explicit resizing operations, and this is the reason why FGSM can be improved by optimizing a single adversarial image.

The experiment also shows that PGD can be damaged by optimizing a single adversarial image. PGD contains 10 gradient ascent steps in our work. \textbf{When optimizing a single adversarial image, PGD has 3$\times$10=30 implicit and 3$\times$10=30 explicit resizing operations, while when using the default implementation introduced in Section 4.1, PGD contains only 3$\times$2=6 explicit resizing operations}. 
In our opinion, the more the number of explicit resizing operations, the worse the GTA performance, and this is the reason why optimizing a single adversarial image at each gradient ascent step damages PGD.

\subsection{Using Knowledge distillation to train a single model}
\label{sec:knowledge distillation}
In this experiment, we use Cifar-10, Tiered$_{T84}$, and Tiered$_{V56}$ as source datasets, and use three RN-18 respectively trained on the source datasets as source models.
First, we build another model $G$ that has three heads that correspond to the three source datasets.
The backbone of $G$ is the same with that of the proposed ICE.
The first head that has 10 output nodes is used to classify Cifar-10 images.
The second head which has 351 output nodes is used to classify Tiered$_{T84}$ images.
The third head which has 257 output nodes is used to classify Tiered$_{V56}$ images.
Second, we use the three source models and three datasets to train a student model $G$ with offline knowledge distillation.
The student model $G$ obtain test accuracies of 97.5\%, 61.7\%, and 70.3\% on Cifar-10, Tiered$_{T84}$, and Tiered$_{V56}$, respectively.
Finally, we use the student model $G$ to disturb the images from Cifar-100.
The GTA success rates on the target models MB-V3, VGG-16, RN-18, and DN-26 are reported in Table~\ref{tab:knowledge}. 
The comparison between the results in Table~\ref{tab:knowledge} with those in the last row of Table 2 (in the main body) indicates that \textbf{attacking a single student model cannot improve the GTA performances.}

\begin{table}[]
	\centering
	\caption{GTA success rates when using a single student model. Target dataset is Cifar-100.
	}
	{\scriptsize  
		\begin{tabular}{ccccc}
			\toprule
			Method & MB-V3 & VGG-16 & RN-18 & DN-26\\
			\midrule
			FGSM & 46.4\% & 59.6\% & 51.0\% & 70.6\% \\  
			PGD &36.9\% & 49.6\% & 40.3\% & 61.3\%  \\
			MI &43.9\% & 56.6\% & 47.6\% & 68.4\%  \\
			\bottomrule
		\end{tabular}
	}
	\label{tab:knowledge}
\end{table}

\section{Computational cost}
We conduct all experiments on Tesla P40 GPU.
The training cost of the proposed ICE is determined by the backbone, the used source datasets, the source models, and \emph{etc.}.
The number of parameters of ICE is determined by the backbone.
The backbone which contains about 7.7M parameters is shown in Figure~\ref{fig:network}.
With the source datasets Cifar-10, Tiered$_{T84}$, and Tiered$_{56}$, and with the three corresponding RN-18 source models, training the model $\mathcal{U}_\theta$ costs about 1.3T FLOPs per iteration when batch size=64.

In inference, the cost of ICE is determined by the backbone, the size of the testing image, and the number of gradient ascent step $T$, which is set to 10 in our work.
The inference costs of PGD-based baselines depend on the size of the testing image, the source models, and the number of gradient ascent steps $T$.
When the testing image comes from Cifar-100, ICE costs approximately 1.2G FLOPs per gradient ascent step per image, while the PGD-based baseline MI, DI, or TI-DIM costs approximately 10.2G FLOPs, which indicates that \textbf{ICE is much more efficient than PGD-based baselines in inference.}

The training cost of the proposed SA is determined by the used source datasets, the source models, and \emph{etc.}.
With the source datasets Cifar-10, Tiered$_{T84}$, and Tiered$_{56}$, and with the three corresponding RN-18 source models, training the parameter $\bm{\omega}$ costs about 0.77T FLOPs per iteration when batch size=64.

In inference, the cost of SA is almost zero because SA perturb the clean image by directly adding the noise $\epsilon \cdot \text{sign}\big(\mathcal{Z}(\bm{\omega}, X_c, Y_c)\big)$ to the clean image.
As a comparison, the baselines RAP and AEG need to firstly generate the adversarial perturbation via generators before adding the perturbation to the clean image.
Similarly, the baseline UAP needs resizing the trained universal perturbation before adding the resized perturbation to the clean image.

In conclusion, in terms of GTA performance, the proposed ICE and SA perform almost comparable and both of them greatly outperform the baselines, and in terms of efficiency, the proposed SA performs the best among all methods because its inference cost is almost zero.

\section{More analysis about SP}
\label{sec:appendix_SP}
The ablation study shown in Section 6.8 demonstrates that the trick SP in this paper is important for ICE and PGD-based baselines to achieve better generalized transferable attack success rates.
It is also observed that without SP, MI performs the best among ICE and all PGD-based baselines.
This is because MI utilizes gradient momentum to improve the attack success rate, and the momentum will enlarge the average perturbation scale of each pixel while keeping the $L_\infty$ of the perturbation map unchanged, which plays a similar role to SP.
For instance, without SP, the average perturbation scale of each pixel of the adversarial examples generated via MI is approximately 11.
As a comparison, for the other PGD-based methods and ICE, the average perturbation scale of each pixel of the generated adversarial examples is no more than 9.

\section{More visualizations of noises and spectrum diagrams}
\begin{figure*}[]
	\centering
	\includegraphics[width=0.7\textwidth]{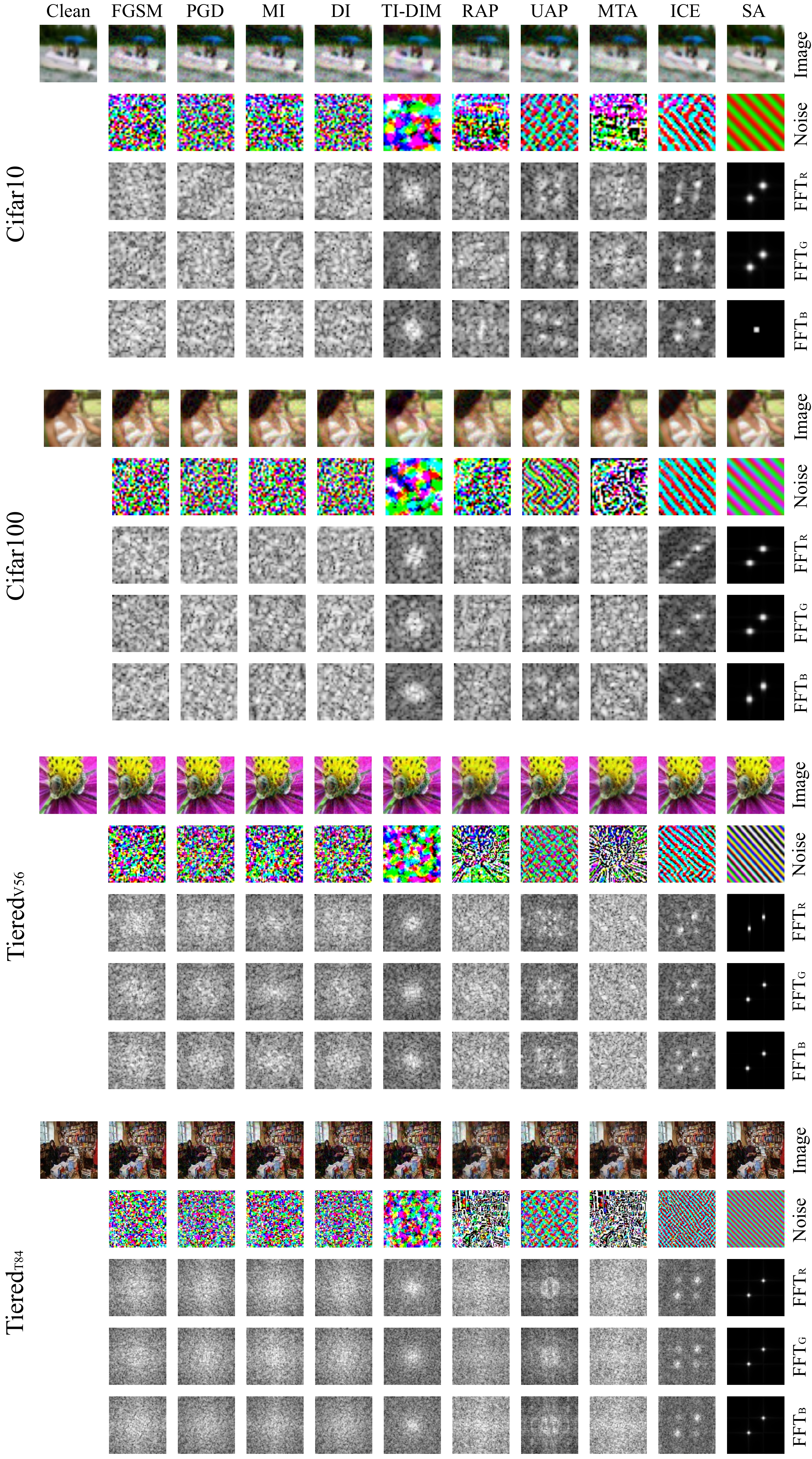}
	\caption{
		Some adversarial examples for clean images from Cifar-10, Cifar-100, Tiered$_{V56}$, and Tiered$_{T84}$.
		The adversarial examples are generated via FGSM, PGD, MI, DI, TI-DIM, RAP, UAP, MTA, ICE, and SA, with $\epsilon=15$.
	}
	\label{fig:visualization_appendix}
\end{figure*}

In Fig.\ref{fig:visualization_appendix}, we visualize more generated adversarial examples, the corresponding noise maps, and spectrum diagrams.
Four datasets Cifar-10, Cifar-100, Tiered$_{V56}$, and Tiered$_{T84}$ are used for the visualization.
For each dataset, we randomly sample a clean image for visualization and use the other three datasets as the source datasets.
For example, when we sample the clean image from Tiered$_{T84}$ for visualization, the source datasets are Cifar-10, Cifar-100, and Tiered$_{V56}$, and the source models are three RN-18 respectively trained on the source datasets.
Each raw FFT$_x$, where $x\in$ [R, G, B], denotes the spectrum diagram of the noise map's $x$ channel.
All the visualizations support the phenomenon that compared with the other methods, ICE noise is more regular, and the main components in ICE noise are three sine functions.
We propose the SA method to directly optimize the three sine waves and further demonstrate that the three sine functions are enough and effective to break DNNs under the GTA setting.
The frequency of each sine wave $Z_j=Sine(a_j\cdot X_{map}+b_j\cdot Y_{map}+c_j)$ is $\sqrt{a_j^2 + b_j^2} / (2\cdot \pi)$, and the wave direction is arctan($b_j/a_j$).
In experiments, we found that the frequencies of the three optimized sine waves are commonly around 0.18HZ, and the directions of the three sine waves are commonly around $45^{\circ}$.

\section{Visualization of perturbed MS-COCO images}
\begin{figure*}[]
	\centering
	\includegraphics[width=0.95\textwidth]{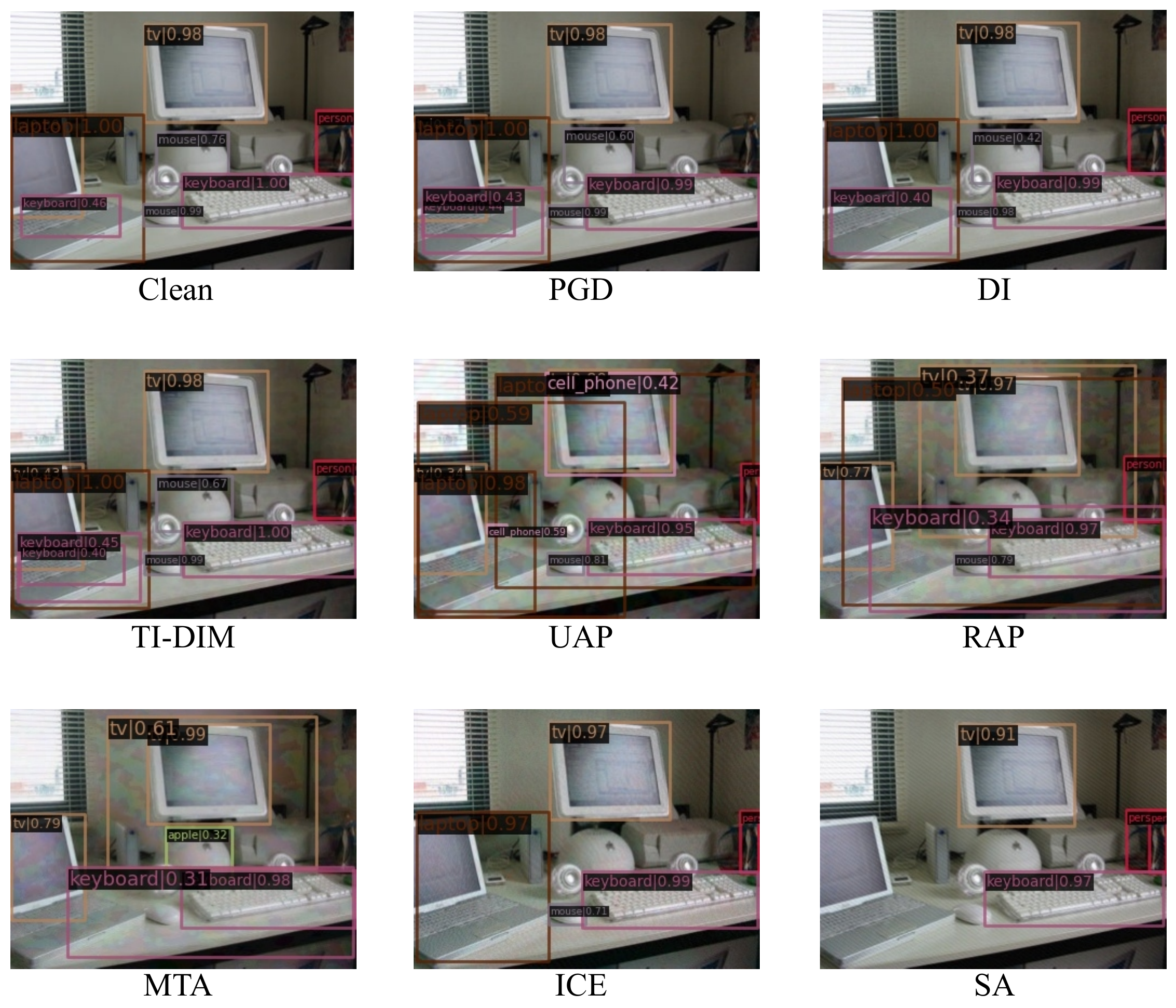}
	\caption{
		An MS-COCO clean image and the corresponding adversarial examples generated via PGD, DI, TI-DIM, UAP, RAP, MTA, ICE, and SA, with $\epsilon=15$.
	}
	\label{fig:visualization_COCO_appendix}
\end{figure*}

Here we visualize a clean MS-COCO image and the corresponding adversarial images generated by PGD, DI, TI-DIM, UAP, RAP, MTA, ICE, and SA, in Fig.\ref{fig:visualization_COCO_appendix}.
The bounding boxes predicted by Faster-RCNN are also visualized for readers to better understand how Faster-RCNN predicts the adversarial examples.
$\epsilon$ is set to 15.
The source datasets are Cifar-10, Tiered$_{T84}$, and Tiered$_{V56}$, and the source models are three ResNet-18 models respectively trained on the three source datasets.
Obviously, among all methods, SA performs the best to perturb MS-COCO images because Faster-RCNN can only detect three objects on the adversarial image generated by SA but can detect more objects on the other adversarial images.

\section{Additional implementation details of baselines}
\label{sec:appendix_baselines}
Some implementation details of baselines have been introduced in Section 6.1 of the main body.
Here we introduce additional implementation details of baselines.

\textbf{MI:} Parameter $\mu$ of MI~\cite{dong2018boosting} is set to 1.

\textbf{{DI:}}.
We use the code\footnote{https://github.com/cihangxie/DI-2-FGSM\label{code}} to implement DI~\cite{xie2019improving} in all our experiments.
We set 'FLAGS.image\_resize' to 36, 64, or 96, when the resolution of the input image is 32, 56, or 84, respectively.
The input diverse possibility $p$ is set to 1.0.

\textbf{TI-DIM:}.
We use the code\footnote{https://github.com/dongyp13/Translation-Invariant-Attacks} to implement TI-DIM~\cite{dong2019evading} in all our experiments.

\textbf{IR:}.
We use the code\footnote{https://github.com/xherdan76/A-Unified-Approach-to-Interpreting-and-Boosting-Adversarial-Transferability} to implement IR~\cite{wu2020skip} in all our experiments.
The hyper-parameter `args.grid\_scale' and `args.sample\_grid\_num' are set to 1 and 16, respectively, for all experiments.

\textbf{FDA+xent:}.
We refer to the FDA+xent paper~\cite{nathan2020perturbing} to implement it on the generalized transferable attack problem.
Given the RN-18 source model, we train $C$ auxiliary binary classifiers based on the third block feature.
$C$ is the number of classes of the RN-18 model.
Each auxiliary classifier contains two hidden layers with each layer containing 100 neurons.
Then, we use the FDA+xent optimizing objective to disturb the auxiliary classifiers and the middle-layer features, and generate perturbation noises for input images following the pipeline introduced in Section 6.1.1.

\textbf{AEG}. We implement AEG~\cite{bose2020adversarial} in our experiment by referring to the code\footnote{https://github.com/joeybose/Adversarial-Example-Games}.
Given each source dataset $D_k$ and the corresponding source models trained on it, we adversarially train a perturbation generator together with a critic.
The generator can be denoted as $G_k$.
For example, for the experiment scene `-Cifar-10', we train three generators on the three source datasets Cifar-100, Tiered$_{T84}$ and Tiered$_{V56}$, and denote them as $G_{1}$, $G_{2}$, $G_{3}$, respectively.
The architecture of all generators is the encoder-decoder defined in Tab.7 of AEG's paper.
Note that considering ground-truth label is unavailable in GTA's inference phase, we do not use the label as the additional input signal for the decoder when training the generators.
Each generator's input-size is the same with the image-shape of the corresponding training dataset.
On either Cifar-10 or Cifar-100, we train the generator and the critic for 500 epochs with the learning rate of 0.001.
On either Tiered$_{T84}$ or Tiered$_{V56}$, we train the generator and the critic for 120 epochs with the learning rate of 0.001.

In the inference phase, we use the following steps to evaluate AEG on generalized transferable attack.
\textbf{1)} For each generator $G_k$, we resize the testing clean image $x$ to the input shapes of the generator, which can be formulated as $x_k=\text{resize}(x, \text{resolution}(G_k) )$, and then feed the resized image $x_k$ to the generator $G_k$. 
\textbf{2)} Obtain the perturbation generated by each generator, which can be formulated as $\delta_k = G_{k}(x_k)$.
\textbf{3)} Average fuse all the perturbations generated by all generators with the formulation $\delta = \frac{1}{m} \sum_{s=1}^{m}\delta_k$, where $m$ is the number of generators.
\textbf{4)} Obtain the adversarial example $\hat{x} = x + \epsilon \cdot \text{sign}(\delta)$, where $\epsilon$ is default set to 15.
\textbf{5)} Feed $x$ and $\hat{x}$ into the unknown target model $\mathbb{M}$, and get the predictions.
\textbf{6)} The generalized transferable attack is successful if $\mathbb{M}(x) \neq \mathbb{M}(\hat{x})$.

\textbf{MTA}.
We refer to the MTA paper~\cite{qin2021training} to implement it on the generalized transferable attack problem.
Given each source dataset $D_k$ and the corresponding source models trained on it, we train a meta-surrogate model, which can be denoted as $\mathcal{S}_k$.
Each meta-surrogate model's input-size is the same with the image-shape of the training dataset.
We train the meta-surrogate models on all the source datasets with the following settings.
On either Cifar-10 or Cifar-100, we train the meta-surrogate model for 50,000 iterations with the parameter $\epsilon_c$ and number of attack steps $T_t$ (in Customized PGD) set to 1600 and 7, respectively.
On either Tiered$_{T84}$ or Tiered$_{V56}$, we train the meta-surrogate model for 70,000 iterations with the parameter $\epsilon_c$ and number of attack steps $T_t$ set to 2100 and 4, respectively.
On each training dataset, $\epsilon_c$ is exponentially decayed by 0.9$\times$ for every 4000 iterations. 
The learning rate and the batch size are set to 0.001 and 64, respectively.

We use the following steps to evaluate MTA on generalized transferable attack.
\textbf{1)} For each meta-surrogate model $\mathcal{S}_k$, we resize the testing clean image $x$ to the input shape of $\mathcal{S}_k$, which can be formulated as $x_k=\text{resize}(x, \text{resolution}(\mathcal{S}_k) )$.
\textbf{2)} We then feed the resized image $x_k$ to the meta-surrogate model $\mathcal{S}_k$.
Because we cannot access the category of the image $x$ in advance, no ground-truth label can be leveraged to perturb the resized image $x_k$.
Therefore, for each meta-surrogate model $\mathcal{S}_k$, we generate adversarial example for $x_k$ by maximizing the entropy (as used in our ICE) for $T$ gradient ascent steps.
The $i$-th step can be formulated as
\begin{equation}
	\left\{
	\begin{array}{lr}
		\hat{y}_{k}^{(i-1)} = \mathcal{S}_{k}(x_{k}^{(i-1)}), \\
		\delta_{k}^{(i-1)} = \text{sign}\big (\nabla_{x_{k}^{(i-1)}}\mathcal{L}(y_{k}^{(i-1)}) ), \\
		x_{k}^{(i)} = \text{clip}(x_{k}^{(i-1)} + \frac{\epsilon}{T} \cdot \delta_{k}^{(i-1)}),
	\end{array}
	\right. 
	\label{eq:generation of the perturbation for MTA}
\end{equation}
where $\hat{y}_{k}^{(i-1)}$ is the meta-surrogate model's output.
$x_{k}^{(i)}$ and
$\delta_{k}^{(i-1)}$ are the adversarial example and the perturbation generated in the $i$-th step, respectively.
\textbf{3)} Resize the adversarial example $x_{k}^{(T)}$ generated by the meta-surrogate model $\mathcal{S}_k$ to the original shape of the image $x$, which can be formulated as $x_{k}' = \text{resize}(x_{k}^{(T)}, \text{resolution}(x))$.
\textbf{4)} Average fuse the adversarial examples generated by all meta-surrogate models to one image $x_{adv}$ following the formulation $x_{adv} = \frac{1}{m} \cdot \sum_{k=1}^{m} \cdot \big(x_{k}' \big)$, where $m$ is the number of meta-surrogate models.
\textbf{5)} Generate adversarial example $\hat{x}$ by the formulation $\hat{x} = \text{clip}(x+\epsilon \cdot \text{sign}(x_{adv} - x))$, which is the SP step defined in Section 4.2 of the main-body. 
\textbf{6)} Feed the adversarial example $\hat{x}$ and the clean image $x$ into the unknown target model $\mathbb{M}$ and get the predictions.
\textbf{7)} The GTA process is successful if $\mathbb{M}(\hat{x}) \neq \mathbb{M}(x)$.

\textbf{UAP}.
The original UAP~\cite{zhang2021data} constructs a universal perturbation within a single dataset.
However, the GTA setting requires the perturbation to be generalized to unknown dataset. 
So we re-implement UAP under the GTA setting by Algorithm~\ref{algorithm:UAP_train}.
We set the initial resolution of the UAP $\bm{\nu}$ to 100$\times$100, and before using it to attack each victim image, we resize it to the resolution of the victim image (See Resize$(\bm{\nu}, \text{resolution}(X_{\mathcal{D}_k}))$ in line 4 of Algorithm~\ref{algorithm:UAP_train}).
The obtained adversarial examples are denoted as $\hat{X}_{\mathcal{D}_k}$.
$\epsilon$, the batch size, and the learning rate $\alpha$ are set to 15, 128, and 0.01, respectively.
We train UAP $\bm{\nu}$ for totally 40,000 iterations.

Given any clean image $x$ that will be fed into an unknown target model $\mathbb{M}$, we evaluate the trained UAP with the following steps.
\textbf{1)} Resize $\bm{\nu}$ to the resolution of $x$.
\textbf{2)} 
Generate the adversarial example $\hat{x}$ by the formulation 
\begin{equation}
	\hat{x} = \text{clip}\Big(x+\epsilon \cdot \text{sign}\big(\text{Resize}(\bm{\nu}, \text{resolution}(x) \ )\big)\Big)
	\label{eq:SA_sign projection}
\end{equation}
\textbf{3)} Feed the adversarial example $\hat{x}$ and the clean image $x$ into the unknown target model $\mathbb{M}$ and get its predictions $\mathbb{M}(\hat{x})$ and $\mathbb{M}(x)$.
\textbf{4)} The GTA process is successful if $\mathbb{M}(\hat{x}) \neq \mathbb{M}(x)$.
\begin{algorithm}[t]
	\caption{Training UAP under the GTA setting}
	\label{algorithm:UAP_train}
	{\bfseries input:} Source datasets $\mathbb{D} \!=\! \{\mathcal{D}_1, \mathcal{D}_2, ...,  \mathcal{D}_m\}$, Source models $\mathbf{M}_{\mathcal{D}_k} \!=\! \{ \mathbf{M}_{\mathcal{D}_k}^1, \mathbf{M}_{\mathcal{D}_k}^2, ..., \mathbf{M}_{\mathcal{D}_k}^{N_k} \}$ for each dataset $\mathcal{D}_k$, Initial perturbation $\bm{\nu}$ with size of 100$\times$100. \\
	{\bfseries output:} Optimized perturbation $\bm{\nu}$. \\
	{\bfseries 1 $\;\!$:} {\bfseries while not} done {\bfseries do} \\
	{\bfseries 2 $\;\!$:} 	\quad	\textbf{for} each $\mathcal{D}_k \in \mathbb{D}$ \textbf{do}\\
	{\bfseries 3 $\;\!$:} 	\qquad	Sample a mini data batch $(X_{\mathcal{D}_k}, Y_{\mathcal{D}_k}) \in \mathcal{D}_k$ \\
	{\bfseries 4 $\;\!$:}   \qquad  $\hat{X}_{\mathcal{D}_k} \!\!=\! \text{Clip}(X_{\mathcal{D}_k} \!+ \epsilon\cdot\text{Resize}(\bm{\nu}, \text{resolution}(X_{\mathcal{D}_k})))$\\
	{\bfseries 5 $\;\!$:}     \quad\quad \textbf{for} each $\mathbf{M}_{\mathcal{D}_k}^j \in \mathbf{M}_{\mathcal{D}_k}$ \textbf{do} \\
	{\bfseries 6 $\;\!$:}     \qquad\quad Obtain loss $\mathbf{L}_{\mathcal{D}_k}^{j}$ on $\hat{X}_{\mathcal{D}_k}$ via Eq.10 of main-body. \\
	{\bfseries 7 $\;\!$:}     \qquad \textbf{end for} \\
	{\bfseries 8 $\;\!$:}     \quad \textbf{end for} \\
	{\bfseries 9 $\;\!$:}  \quad   $\bm{\nu} \!=\! \bm{\nu} \!+\! \alpha\!\cdot\!\nabla_{\bm{\nu}} \big( \frac{1}{m}\sum_{k=1}^{m} (\frac{1}{N_k} \sum_{j=1}^{N_k} \mathbf{L}_{\mathcal{D}_k}^{j}) \big)$ \\
	{\bfseries 10:} 	{\bfseries end while} \\
	{\bfseries 11:} 	{\bfseries return} ${\bm{\nu}}$
\end{algorithm}

\textbf{RAP}. We implement RAP~\cite{naseer2019cross-domain} in our work by referring to the official paper and code\footnote{https://github.com/Muzammal-Naseer/Cross-Domain-Perturbations}.
Given each source dataset $D_k$ and the corresponding source models trained on it, we train a perturbation generator following Algorithm 1 in RAP's official paper.
The generator can be denoted as $G_k$.
For example, for the experiment scene `-Cifar-10', we train three generators on the three source datasets Cifar-100, Tiered$_{T84}$ and Tiered$_{V56}$, and denote them as $G_{1}$, $G_{2}$, $G_{3}$, respectively.
The architecture of all generators is the encoder-decoder defined in RAP's official code (`generators.py' file).
Each generator's input-size is the same with the image-shape of the corresponding training dataset.
After training the generators, we use the evaluation pipeline of AEG to evaluate RAP under the GTA setting.

\end{document}